\documentclass{article} % For LaTeX2e
\usepackage{iclr2027_conference,times}

\usepackage{amsmath,amsfonts,bm}

\def\eqref#1{equation~\ref{#1}}
\def\1{\bm{1}}

\DeclareMathAlphabet{\mathsfit}{\encodingdefault}{\sfdefault}{m}{sl}
\SetMathAlphabet{\mathsfit}{bold}{\encodingdefault}{\sfdefault}{bx}{n}

\def\cJ{{\mathcal{J}}}

\def\cP{{\mathcal{P}}}

\def\cV{{\mathcal{V}}}

\newcommand{\Var}{\mathrm{Var}}

\newcommand{\TheName}{FReDraft}

\usepackage{hyperref}
\usepackage{url}
\usepackage{booktabs, makecell, multirow, tabularx, graphicx}
\usepackage{threeparttable}
\usepackage{amsthm}
\usepackage{enumitem}
\usepackage{algorithm, algorithmic}
\usepackage{subcaption}
\usepackage{xcolor}
\usepackage{color}
\usepackage{amsthm}
\usepackage{float}
\usepackage{wrapfig}
\usepackage{tikz}
\usepackage{tcolorbox}
\usepackage{amssymb}
\usepackage{xcolor}
\usepackage{times}
\usepackage{latexsym}
\usepackage[T1]{fontenc}
\usepackage{tcolorbox}
\usepackage[utf8]{inputenc}
\usepackage{microtype}
\usepackage{bbm}
\usepackage{graphicx}
\usepackage{microtype}
\usepackage{colortbl}

\newtheorem{theorem}{Theorem}
\newtheorem{corollary}[theorem]{Corollary}

\usepackage{booktabs}
\usepackage{pifont}

\usepackage[normalem]{ulem}
\newtheorem{definition}{Definition}

\newtheorem{proposition}{Proposition}

\title{Directly Maximizing On-policy Expected Accepted Length for Speculative Decoding}

\author{Ziying Zhang$^{1,2}$
	\qquad Yaqing Wang$^{3}$
	\qquad Yong Liu$^{2,4}$
	\qquad Quanming Yao$^{1,2}$\\
	$^1$Department of Electronic Engineering, Tsinghua University \qquad
	$^2$Zhongguancun Academy\\
	$^3$BIMSA \qquad
	$^4$Gaoling School of Artificial Intelligence, Renmin University\\
	\texttt{ziying-z21@mails.tsinghua.edu.cn, qyaoaa@tsinghua.edu.cn} \\
}

\iclrfinalcopy % Uncomment for camera-ready version, but NOT for submission.
\begin{document}

\maketitle

\begin{abstract}
Speculative decoding accelerates large language model inference by using a lightweight draft model to propose tokens for verification by a target model.
Draft training commonly optimizes token-level distribution alignment, but this does not directly optimize the length of the accepted prefix: once a draft token is rejected, all subsequent draft tokens are discarded.
We propose \TheName{}, an on-policy training method that directly maximizes the expected accepted length of drafts sampled from the current draft model.
This objective accounts for first-rejection truncation and induces prefix-aware credit assignment, weighting each token by its contribution to the acceptance of the current and subsequent tokens.
To optimize the objective efficiently, we analytically marginalize token-level acceptance rate, using the target model's vocabulary distribution to reduce gradient variance.
We further redistribute part of this distribution-level gradient into an explicit penalty for sampled tokens whose probabilities are overestimated by the draft model, while preserving the expected gradient.
Experiments across multiple benchmarks and target models show that \TheName{} improves accepted length and decoding speedup over strong baselines, achieving faster on-policy training convergence.
\end{abstract}

\section{Introduction}\label{sec:intro}

\begin{figure}[t]
     \vspace{-4pt}
    \centering
    \includegraphics[width=\linewidth]{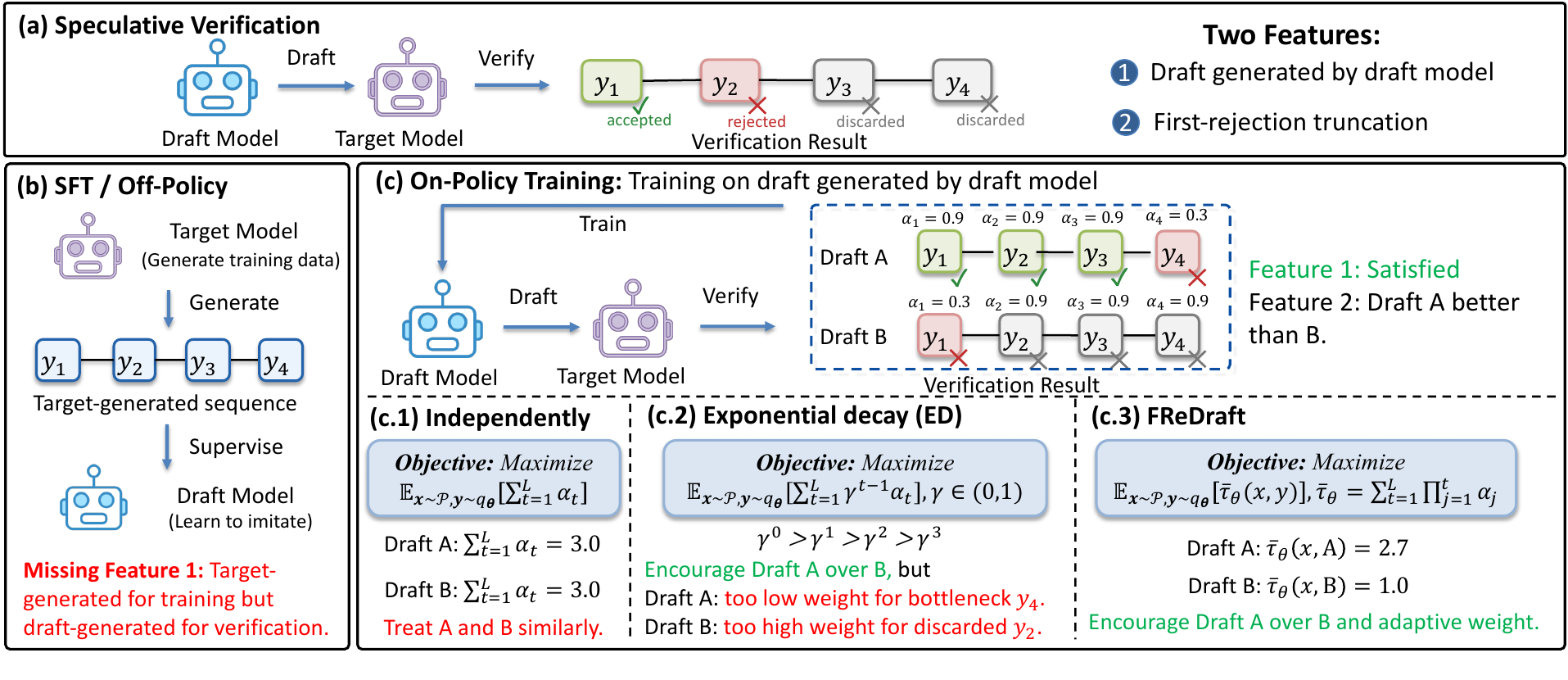}
    \vspace{-24pt}
	\caption{\label{fig:motivation}
    \TheName{} trains on self-generated drafts and accounts for first-rejection truncation. Here $\alpha$ means token-level acceptance rate and $\bar{\tau}$ means expected accepted length.
    }
    \vspace{-20pt}
\end{figure}

Large language models (LLMs) have achieved strong performance across a wide range of generation tasks, including reasoning, coding and agentic tasks~\citep{r1,qwen3,k2.5}.
But their autoregressive decoding process remains time-consuming because each new token requires one LLM forward pass~\citep{pope2023efficiently,respec}.
Speculative decoding accelerates LLM inference through a draft-verification paradigm (Figure~\ref{fig:motivation} (a)): a lightweight draft model proposes multiple future tokens, and the target LLM verifies them in one forward pass with a rejection-sampling rule that preserves the target distribution~\citep{sd1,sd2}.
Since acceleration comes from accepting multiple draft tokens per target-model call, the speedup depends heavily on the acceptance rate of tokens in the draft~\citep{zhang2025speculative,qspec}.

To train the draft model for higher acceptance rate, a typical approach is supervised finetuning (SFT) on target-generated sequences~\citep{eagle3,dflash,specforge}, minimizing token-level cross-entropy (CE) loss independently.
However, token verification within a draft is not independent; instead, it exhibits ``first-rejection truncation'' property~\citep{griffin,vat}.
Verification proceeds sequentially from the first draft token, and once a token is rejected, all subsequent tokens are discarded.
Consequently, a later token contributes to speedup only if its entire prefix survives verification.
Inspired by this, some works explore verification-aware draft model training~\citep{pard2}.
For example, LK-Losses~\citep{lk} maximizes exponential decay (ED) weighted token-level log acceptance rate, VSD~\citep{vsd} optimizes a variational objective as lower bound of draft tree utility, whereas D-PACE~\citep{dpace} proposes to maximize the summed cumulative draft confidence at each position in target-generated sequences.

These verification-aware methods achieve draft model enhancement over naive SFT, but they actually optimize a metric of sequences under target-distribution as a surrogate, which does not match ``first-rejection truncation'' in practical verification (Figure~\ref{fig:motivation} (b)).
First, draft model is trained on sequences following target distribution (or hybrid of draft and target distribution), whereas evaluated on drafts proposed under its own policy, leading to exposure bias.
Second, the optimized metrics do not necessarily mean later ``first-rejection''.
For example, optimizing cumulative draft confidence makes draft model diverge from target severely when target model itself is uncertain (See Appendix~\ref{app:dpace} for systematic analysis).
Recently, Draft-OPD~\citep{draftopd} proposes on-policy paradigm to train the draft model on its own generated drafts, eliminating exposure bias~\citep{gto} (Figure~\ref{fig:motivation} (c)).
However, it applies heuristically designed forward and reverse KL weighted by static ED.
Such static schedules cannot capture actual importance of tokens, which depends on decoding context, and target-to-draft probability ratios encountered during verification.
As a result, a token constituting the verification bottleneck may receive insufficient weight, whereas tokens following prefixes with low survival probability may be overemphasized, leading to inefficient allocation of the training signal.

Draft accepted length is a metric directly related to speedup ratio of speculative decoding~\citep{magicdec,zhang2025speculative,pard2}.
Given a draft generated by draft model, its expected accepted length can be calculated precisely by summing the prefix survival rate of token at each position.
Therefore, this metric intrinsically contains influence of verification result from earlier tokens to later ones (Figure~\ref{fig:motivation} (c.2)), matching first-rejection truncation property~\citep{gto,opt-tree}.
Inspired by the advantage of expected accepted length, we propose \textbf{\TheName{}}, a draft model training method that directly maximizes the expected accepted length of draft generated by draft model itself, to model first-rejection truncation.
The objective naturally matches on-policy paradigm, and derived gradient captures prefix survival and influence of earlier tokens to later tokens in practical verification.
But directly optimizing draft model with this gradient remains inefficient.
In practice, the gradient requires Monte Carlo (MC) sampling to estimate, from which we identify two factors that make optimization inefficient:
First, only target model probabilities for sampled draft tokens can be used to estimate the gradient, leading to high gradient estimation variance and the vocabulary distribution output by target model during verification discarded.
Second, over-sampled tokens, for which the draft model assigns higher probability than target model, receive zero gradient.
Since on-policy training data is generated by the draft model itself, such over-sampled tokens are common and become ineffective in training.
To tackle these difficulties, \TheName{} proposes to marginalize single token acceptance rate expectation, which reduces the variance of gradient estimation and enables leveraging of target model vocabulary distribution to supervise draft model training.
\TheName{} also introduces penalty for over-sampled tokens, allowing frequently-appearing over-sampled tokens to receive non-zero gradient.
We summarize the contributions as follows:
\begin{itemize}[leftmargin=*]
\item We formulate draft training as on-policy maximization of expected accepted length, aligning the objective with first-rejection truncation and deriving token-level credit assignment from each token's contribution to prefix survival.

\item We develop token-level acceptance marginalization to reduce gradient variance with target vocabulary distribution and redistribute part of the resulting gradient into an explicit over-sampled token penalty, enabling efficient optimization while preserving the expected gradient.

\item Experiments across multiple target models and benchmarks show that \TheName{} improves accepted draft length and decoding speedup over strong baselines, including Draft-OPD.
It also exhibits faster convergence during on-policy training.
\end{itemize}

\section{\TheName}\label{sec:method}
\begin{figure}[t]
%\vspace{-1.2em}
\centering
\includegraphics[width=\linewidth]{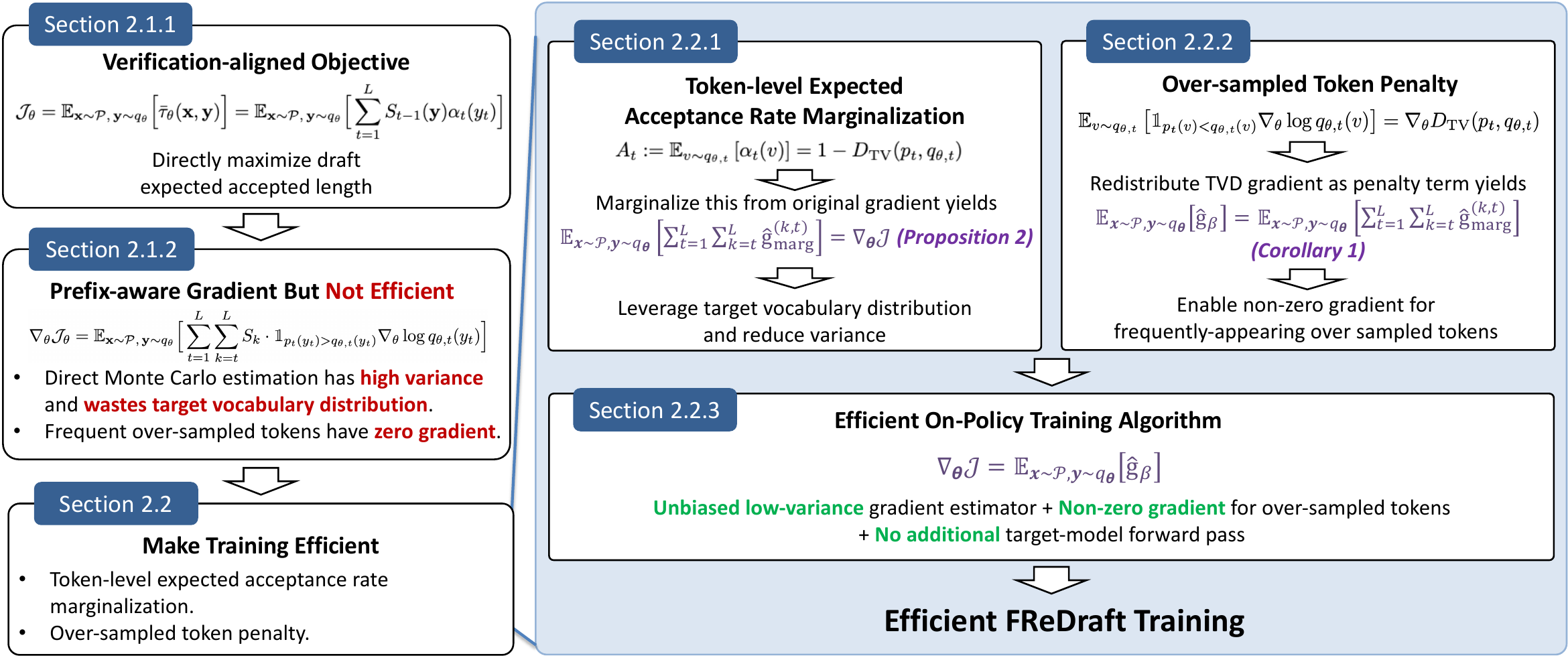}
\caption{Overview of how \TheName{} preserves the verification-aligned objective while constructing an efficient on-policy gradient estimator.}
\label{fig:optimization-overview}
\vspace{-1.0em}
\end{figure}

Section~\ref{sec:method} introduces \TheName{}, a draft model on-policy training method that maximizes self-generated drafts' expected accepted length, enabling first-rejection-truncation-aware training signal allocation.
Section~\ref{sec:app-pf-problem} formulates the expected accepted length objective, characterizes its gradient and identifies two factors resulting in optimization inefficiency.
To deal with the two factors, Section~\ref{sec:optimization} develops optimization techniques for efficient training.
We show the overall outline of Section~\ref{sec:method} in Figure~\ref{fig:optimization-overview}.

\subsection{Verification-Aligned Objective}\label{sec:app-pf-problem}

\subsubsection{Objective Formulation}\label{sec:app-pf-formulation}

Let $p$ be the fixed target model and $q_\theta$ the trainable draft model over vocabulary $\mathcal{V}$,
given a decoding context $\mathbf{x}\sim\cP$ (referring to Draft-OPD~\citep{draftopd}, $\cP$ is the distribution of decoding context that will be visited by $p$ and $q_{\theta}$ when conducting speculative decoding), speculative decoding first employs $q_{\theta}$ to propose a length-$L$ ($L>1$) draft block $\mathbf{y}=(y_1,\ldots,y_L)\sim q_\theta$.
Define $h_t=(\mathbf{x},\mathbf{y}_{<t})$ and
$p_t(v)=p(v\mid h_t)$, $q_{\theta,t}(v)=q_\theta(v\mid h_t)$ ($q_{\theta,t}(v)=q_\theta(v\mid \mathbf{x})$ for parallel draft model e.g. DFlash~\citep{dflash}).
Then during verification, token $y_t$ is accepted with probability
\begin{equation}\label{eq:app-pf-acceptance}
	\alpha_t(y_t) = \min \big\{1, p_t(y_t) / q_{\theta,t}(y_t)\big\}.
\end{equation}
Verification stops at the first rejection, so it evaluates an accepted prefix rather than independent token predictions~\citep{sd1,sd2}.
Consequently, token $y_t$ contributes only when every preceding token survives.
To model such first-rejection truncation property of verification, we learn $q_\theta$ by directly maximizing the expected accepted length of the drafts generated by $q_\theta$.

\begin{definition}[Verification-aligned expected accepted-length objective]
	\label{def:app-pf-objective}
	Given a decoding context $\mathbf{x}\sim\cP$ and a length-$L$ draft
	$\mathbf{y}=(y_1,\ldots,y_L)\sim q_\theta$, the
	verification-aligned objective is
	\begin{equation}
		\cJ_{\theta} =
		\mathbb{E}_{\mathbf{x}\sim\cP,\,
			\mathbf{y}\sim q_{\theta}}
		\Big[
		\bar{\tau}_{\theta}(\mathbf{x},\mathbf{y})
		\Big] =
        \mathbb{E}_{\mathbf{x}\sim\cP,\,
			\mathbf{y}\sim q_{\theta}}
        \Big[\sum\nolimits_{t=1}^{L}S_{t-1}(\mathbf{y})\alpha_t(y_t)\Big]
        ,
		\label{eq:app-pf-objective}
	\end{equation}
	where
	$\bar{\tau}_{\theta}(\mathbf{x},\mathbf{y})
	=
	\sum_{t=1}^{L}\prod_{j=1}^{t}\alpha_j(y_j)$
	is the expected accepted length of $\mathbf{y}$ (see Appendix~\ref{app:expected_al} for derivation), and $S_{t-1}(\mathbf{y}) = \prod_{j=1}^{t-1}\alpha_j(y_j)$, with $S_0(\mathbf{y})=1$, is the probability that verification reaches position $t$ (i.e. prefix survival rate of prefix $\mathbf{y}_{<t}$).
\end{definition}

Definition~\ref{def:app-pf-objective} defines the expected accepted length maximization objective and further characterizes it as token-level acceptance rate maximization with prefix-survival $S_{t-1}$ as the coefficient, rather than treating all token positions independently.
This objective aligns with verification process of speculative decoding in terms of following two aspects:
(1) The verified drafts are generated by the draft model $q_\theta$ itself, matching the real-world scenario and making the objective naturally suitable for on-policy training. (2) The maximized $\bar{\tau}_{\theta}(\mathbf{x},\mathbf{y})$ intrinsically captures the influence of earlier token verification result to later tokens, thus matching the first-rejection truncation property of verification.

\subsubsection{Gradient Theoretical Characterization}
\label{sec:app-pf-characterization}

Definition~\ref{def:app-pf-objective} provides a verification-aligned objective based on expected accepted length.
We next identify the optimization direction induced from the objective by analyzing its gradient.

\begin{proposition}[Prefix-aware gradient]
\label{prop:app-pf-gradient}
Let $G_t = \mathbbm{1}_{p_t(y_t)>q_{\theta,t}(y_t)} \nabla_{\theta} \log q_{\theta,t}(y_t)$ be the gradient for token $y_t$ at position $t$ and $\widehat{\mathbf{g}}_{\mathrm{dir}}^{(k,t)} = S_kG_t$.
The gradient of the objective in Equation~\ref{eq:app-pf-objective} is
\begin{equation}
	\nabla_{\theta}\cJ_{\theta} 
    = \mathbb{E}_{\mathbf{x}\sim\cP,\, \mathbf{y}\sim q_{\theta}}\Big[\sum\nolimits_{t=1}^{L}\sum\nolimits_{k=t}^{L}\widehat{\mathbf{g}}_{\mathrm{dir}}^{(k,t)}\Big] = \mathbb{E}_{\mathbf{x}\sim\cP,\, \mathbf{y}\sim q_{\theta}}\Big[\sum\nolimits_{t=1}^{L}\Big(\sum\nolimits_{k=t}^{L}S_k\Big)G_t\Big].
	\label{eq:app-pf-gradient}
\end{equation}
\end{proposition}

In practice, $\cP$ depends on $q_{\theta}$ but $\nabla_{\theta}\cP$ is too hard to estimate.
Therefore, referring to Draft-OPD~\citep{draftopd}, Equation~\ref{eq:app-pf-gradient} omits $\nabla_{\theta}\cP$, and what it actually optimizes is ``expected accepted length of draft-model-generated draft for given decoding context $\mathbf{x}$''
(See Appendix~\ref{app:direct-gradient} for derivation).
Proposition~\ref{prop:app-pf-gradient} shows that the objective induces prefix-aware credit assignment.
The weight $\sum_{k=t}^{L}S_k(\mathbf{y})$ for $G_t$ aggregates the survival probabilities of all prefixes extending through position $t$ whose survival depends on token $y_t$.
Therefore, the optimization weight of a token reflects not only its own acceptance but also its influence on subsequent verification outcomes, matching the first-rejection mechanism.

However, directly training the draft model as Equation~\ref{eq:app-pf-gradient} remains inefficient, where we identify two key factors.
\textbf{First}, the trajectory-dependent weight $\sum_{k=t}^{L}S_k(\mathbf{y})$ must be estimated through MC sampling and only target probabilities on sampled tokens can be used to estimate it.
This leads to high variance and the target vocabulary distribution discarded during on-policy training.
\textbf{Second}, over-sampled tokens satisfying $p_t(y_t)<q_{\theta,t}(y_t)$ receive zero gradient due to the indicator $\mathbbm{1}_{p_t(y_t)>q_{\theta,t}(y_t)}$.
Moreover, such tokens frequently appear in training because the training data is generated by $q_{\theta}$.

% ----------------------

\subsection{Efficient Optimization}\label{sec:optimization}

To deal with above two factors, as summarized in Figure~\ref{fig:optimization-overview}, we develop two techniques, i.e. token-level acceptance rate marginalization (Section~\ref{sec:decomposition}) and over-sampled-token penalty (Section~\ref{sec:penalty}), to make the optimization more efficient, while preserving the expected gradient of original objective.

\subsubsection{Token-level Expected Acceptance Rate Marginalization}
\label{sec:decomposition}

The expectation of gradient in Equation~\ref{eq:app-pf-gradient} requires Monte Carlo sampling to estimate, introducing high variance.
It also leads to that only target probabilities for sampled tokens can be used to estimate the gradient, leaving target vocabulary distribution discarded.
Actually, the token-level expected acceptance rate can be computed precisely with no need for MC sampling.
Specifically, for a fixed decoding context $\mathbf{x}$ and draft prefix $\mathbf{y}_{<t}$, 
the expected acceptance rate of the next token is given by
\begin{equation}
    A_t
    :=
    \mathbb{E}_{v\sim q_{\theta,t}}
    \left[\alpha_t(v)\right]
    =
    \sum\nolimits_{v\in\mathcal{V}}
    \min\left\{p_t(v),q_{\theta,t}(v)\right\}
    =
    1-D_{\mathrm{TV}}(p_t,q_{\theta,t}).
\label{eq:acceptance-overlap}
\end{equation}
Here $D_{\mathrm{TV}}(p_t,q_{\theta,t}) = \frac{1}{2}\sum_{v\in \cV}|p_t(v) - q_{\theta,t}(v)|$ is total variation distance (TVD)~\citep{Handel2014ProbabilityIH}.
Unlike sampled acceptance event, $A_t$ can be computed exactly according to vocabulary distribution of draft and target model, produced during verification.
Marginalizing this expected acceptance rate from Equation~\ref{eq:app-pf-gradient} transforms part of the Monte-Carlo estimated gradient into precise computation, thus yielding a lower-variance estimator of the gradient as follows (See Appendix~\ref{app:gradient1} for full derivation):

\begin{proposition}[Expected acceptance rate marginalization]
\label{prop:marginalization}
Define the gradient estimator
\begin{equation}
\begin{aligned}
    \widehat{\mathbf{g}}_{\mathrm{marg}}^{(k,t)}
    := 
    \begin{cases}
        -S_{t-1}(\mathbf{y})\nabla_{\theta}D_{\mathrm{TV}}(p_t,q_{\theta,t}), & k = t, \\
        A_kS_{k-1}G_t, & k>t.
    \end{cases}
\end{aligned}
\label{eq:gradient1}
\end{equation}
Then
$\mathbb{E}_{\mathbf{x}\sim\cP, \mathbf{y}\sim q_{\theta}}[\sum\nolimits_{t=1}^{L}\sum\nolimits_{k=t}^{L}\widehat{\mathbf{g}}_{\mathrm{marg}}^{(k,t)}]
= \mathbb{E}_{\mathbf{x}\sim\cP, \mathbf{y}\sim q_{\theta}}[\sum\nolimits_{t=1}^{L}\sum\nolimits_{k=t}^{L}\widehat{\mathbf{g}}_{\mathrm{dir}}^{(k,t)}]
= \nabla_{\theta}\cJ_{\theta}$.
Under the same Monte Carlo sampling scheme and any fixed direction $\mathbf{a}$ of compatible dimension, $\Var(\mathbf{a}^{\top}\widehat{\mathbf{g}}_{\mathrm{marg}}^{(k,t)} | \mathbf{x}, \mathbf{y}_{<t}) \le \Var(\mathbf{a}^{\top}\widehat{\mathbf{g}}_{\mathrm{dir}}^{(k,t)}| \mathbf{x}, \mathbf{y}_{<t}), \forall 1 \le t \le k \le L$, where $\Var(\cdot)$ means variance.
\end{proposition}

Proposition~\ref{prop:marginalization} analytically marginalizes the terminal-token sampling in each current and future prefix-credit contribution while retaining trajectory-dependent credit assignment through the draft prefix.
Equation~\ref{eq:gradient1} consists of two terms: 
The term with $G_t$ is gradient for sampled token $y_t$, while the term with $\nabla_{\theta}D_{\mathrm{TV}}(p_t,q_{\theta,t})$ is gradient over the vocabulary.
Equation~\ref{eq:gradient1} leverages the target vocabulary distribution available during verification to compute $\nabla_{\theta}D_{\mathrm{TV}}(p_t,q_{\theta,t})$.
Compared with KL, the gradient of TVD w.r.t logits does not shrink when two distributions are close, which has been associated with better optimization efficiency in this case~\citep{ji2023tailoring,lk} (See Appendix~\ref{app:penalty} for detailed discussion).
Notably, variance reduction is for $\widehat{\mathbf{g}}_{\mathrm{marg}}^{(k,t)}$ rather than the summed estimator of $\nabla_{\theta}\cJ_{\theta}$, which also depends on covariance among draft positions and difficult to characterize theoretically.
We provide empirical evidence for total variance reduction in Section~\ref{sec:in-depth}.

\subsubsection{Over-Sampled Token Penalty}
\label{sec:penalty}

Token-level expected acceptance rate marginalization introduces dense distribution supervision over the vocabulary, but the gradient term for sampled tokens in Equation~\ref{eq:gradient1} remains gated by $\mathbbm{1}_{p_t(y_t)>q_{\theta,t}(y_t)}$.
Consequently, over-sampled tokens satisfying $q_{\theta,t}(y_t)>p_t(y_t)$ receive no direct sampled-token correction, although they occur frequently under on-policy sampling.
One possible remedy is sampling from a mixture of the target and draft distributions to reduce the ratio of over-sampled tokens and apply importance weighting~\citep{minillm} for unbiased estimation.
However, this requires additional auto-regressive sampling from the target model and undermines efficiency.
Instead, we redistribute the dense TVD gradient $\nabla_{\theta}D_{\mathrm{TV}}(p_t,q_{\theta,t})$ available in Equation~\ref{eq:gradient1}.
Specifically, define $\widetilde{G}_t = \mathbbm{1}_{p_t(y_t) < q_{\theta,t}(y_t)}\nabla_{\theta}\log q_{\theta,t}(y_t)$,
then almost everywhere, it admits:
\begin{equation}
    \mathbb{E}_{y_t \sim q_{\theta,t}}
    [
        \widetilde{G}_t
    ]
    =
    \nabla_{\theta}D_{\mathrm{TV}}(p_t,q_{\theta,t}).
\label{eq:sampled_tv}
\end{equation}
The two sides of Equation~\ref{eq:sampled_tv} have the same conditional expectation given the decoding context $\mathbf{x}$ and draft prefix $\mathbf{y}_{<t}$.
We can therefore replace a controlled fraction of the dense TVD term by its sampled representation without changing the expected gradient as follows (see Appendix~\ref{app:gradient2} for full proof):
\begin{corollary}[Expectation-preserving over-sampled-token correction]
\label{cor:hybrid}
For any $\beta \in [0,1]$, define
\begin{equation}
    \widehat{\mathbf{g}}_{\beta}
    :=
    \sum\nolimits_{t=1}^{L}\sum\nolimits_{k=t}^{L}\widehat{\mathbf{g}}_{\mathrm{marg}}^{(k,t)}
    +
    \beta \sum\nolimits_{t=1}^{L}S_{t-1}(\mathbf{y})
    \Big[
        \nabla_{\theta}D_{\mathrm{TV}}(p_t,q_{\theta,t}) 
        -
        \widetilde{G}_t
    \Big].
\label{eq:gradient2}
\end{equation}
Then
$\mathbb{E}_{\mathbf{x}\sim\cP, \mathbf{y}\sim q_{\theta}}[\widehat{\mathbf{g}}_{\beta}]
=
\mathbb{E}_{\mathbf{x}\sim\cP, \mathbf{y}\sim q_{\theta}}[\sum\nolimits_{t=1}^{L}\sum\nolimits_{k=t}^{L}\widehat{\mathbf{g}}_{\mathrm{marg}}^{(k,t)}]
=
\nabla_{\theta}\cJ_{\theta}$.
\end{corollary}

Corollary~\ref{cor:hybrid} redistributes existing, rather than introduces additional, gradient information.
Expanding Equation~\ref{eq:gradient2} gives the position weight
$w_t^{(\beta)}(\mathbf{y})
=
[\sum_{k=t+1}^{L}A_kS_{k-1}(\mathbf{y})]
\mathbbm{1}_{p_t(y_t)>q_{\theta,t}(y_t)}
-
[\beta S_{t-1}(\mathbf{y})]
\mathbbm{1}_{p_t(y_t) < q_{\theta,t}(y_t)}$ for gradient of sampled token $\nabla_{\theta}\log q_{\theta,t}(y_t)$.
Intuitively, such a weight encourages the under-sampled tokens with $p_t(y_t)>q_{\theta,t}(y_t)$ while penalizing over-sampled tokens with $p_t(y_t) < q_{\theta,t}(y_t)$.
The gradient of dense TVD term $\nabla_{\theta}D_{\mathrm{TV}}(p_t,q_{\theta,t})$ is weighted by $-(1-\beta)S_{t-1}(\mathbf{y})$.
Thus, $\beta=0$ recovers Equation~\ref{eq:gradient1}, whereas increasing $\beta$ transfers more of the dense TVD gradient to explicit sampled-token penalty for over-sampled tokens, enabling them to receive non-zero gradient.
This transformation does not introduce additional information for training.
However, it weakens the stochastic optimization fluctuation of long-tailed unsampled tokens coupled with vocabulary TVD, but preserves the gradient of over-sampled tokens, which are usually bottleneck for draft accepted length.
$\beta$ works here as a hyperparameter to balance the gradient for over-sampled tokens and vocabulary.
See Appendix~\ref{app:penalty} for detailed discussion about effectiveness of penalty term.

\subsubsection{On-Policy Optimization}\label{sec:algorithm}

\begin{wrapfigure}{r}{0.53\columnwidth}
\vspace{-20pt}
	\begin{minipage}{0.53\textwidth}
\begin{algorithm}[H]
    \small
    \caption{\label{alg:training}On-Policy Optimization with \TheName{}}
\begin{algorithmic}[1]
	\REQUIRE Prompt set $\cP_p$, frozen target model $p$, initialized draft model $q_{\theta}$, coefficient $\beta$, learning rate $\eta$
	\ENSURE Optimized draft model $q_{\theta}$
	
	\WHILE{\textnormal{not converged}}
	\STATE Sample a prompt batch from $\cP_p$
	\STATE Run speculative decoding with $q_{\theta}$ and the frozen $p$
	\STATE Collect drafts and their target/draft distributions
	\STATE Compute the batch estimator $\widehat{\mathbf{g}}_{\beta}$ with Equation~\ref{eq:gradient2}
	\STATE $\theta\leftarrow
	\theta+\eta\widehat{\mathbf{g}}_{\beta}$
	\ENDWHILE
	\STATE \textbf{return} $q_{\theta}$
\end{algorithmic}
\end{algorithm}
	\end{minipage}
	\vspace{-15pt}
\end{wrapfigure}

The objective in Equation~\ref{eq:app-pf-objective} is inherently on-policy because its expectation is taken over drafts generated by the current draft model $q_{\theta}$. 
Specifically, for a batch of prompts sampled from training dataset, we conduct the standard speculative decoding for them, during which we collect a series of draft blocks $\mathbf{y}$ for each prompt, and corresponding target model distribution $p$ and draft model distribution $q_{\theta}$.
Then we estimate the gradient with the collected information according to Equation~\ref{eq:gradient2}.
\TheName{} reuses the distributions already computed during verification and requires neither auto-regressive target sampling nor additional target forward pass.
For $\nabla_{\theta}D_{\mathrm{TV}}$, we adopt the widely used top-K target tokens in OPD to estimate it to save memory in practice
(Theoretical and empirical results show that top-K introduces negligible error compared with full vocabulary. See Appendix~\ref{app:topk} for details)~\citep{fu2026revisiting}.
Compared with existing on-policy method (e.g. Draft-OPD), \TheName{} requires no extra target model forward pass and achieves comparable per-training-step efficiency (See Appendix~\ref{app:wall-clock} for wall-clock time comparison).
We summarize the proposed on-policy algorithm in Algorithm~\ref{alg:training}.
To make the on-policy training more efficient and stable, we refer to Draft-OPD~\citep{draftopd} and first warm up the draft model with SFT on target-generated sequences.
Then we further finetune it as Algorithm~\ref{alg:training}.

\section{Comparison with Representative Draft Training Methods}\label{sec:comparison}

To position \TheName{}, Table~\ref{tab:method_comparison} compares it with representative draft-model training methods.
Naive SFT~\citep{specforge,eagle3,dflash} and off-policy verification-aware method (e.g. LK-Losses~\citep{lk} and D-PACE~\citep{dpace}) train the draft model on target-generated sequences.
The difference is that verification-aware methods introduce verification information into training objective.
For example, LK-Losses optimizes token-level log acceptance rate, and D-PACE maximizes summed cumulative draft confidence on target-generated sequences (see appendix~\ref{app:dpace} for more detailed discussion about D-PACE).
However, objective of these methods does not come from practical verification and cannot necessarily enable later ``first-rejection'' in practical verification, due to exposure bias, and the gap between optimized surrogate and draft accepted length.
Draft-OPD~\citep{draftopd} adopts on-policy distillation~\citep{li2026rethinking} to mitigate exposure bias.
But it applies ED to weight token-level KL, which can lead to inefficient training signal allocation.
In contrast, \TheName{} directly maximizes the expected accepted length of drafts generated by the draft model, achieving on-policy training.
The derived credit assignment evaluates prefix survival and influence on subsequent verification outcomes in practical verification, matching the first-rejection truncation property of verification process.
See Appendix~\ref{app:related-works} for detailed discussion about more related works.
\begin{table}[htbp]
	\centering
	\small
	\setlength{\tabcolsep}{5pt}
    \vspace{-5px}
	\caption{Comparison of draft-model training methods. V-PS and V-SO denote practical-Verification-defined Prefix-Survival and influence on Subsequent verification Outcomes in credit assignment.}
	\vspace{-5px}
	\label{tab:method_comparison}
	\begin{tabular}{lccccc}
		\toprule
		\textbf{Method}
		& \textbf{Objective} & \textbf{Data} & \textbf{Credit} & \textbf{V-PS} & \textbf{V-SO} \\
		\midrule 
        SFT & 
        Token-level cross entropy & 
        Off-policy & 
        Uniform / ED  & 
        $\times$ &
        $\times$
        \\\midrule
		D-PACE & 
        Summed cumulative draft confidence & 
        Off-policy & 
        Adaptive & 
        $\times$ &
        $\times$
        \\
		LK-Losses & 
        Token-level log acceptance rate & 
        Off-policy & 
        ED  & 
        $\times$&
        $\times$
        \\\midrule
		Draft-OPD & 
        Forward / Reverse KL divergence & 
        On-policy & 
        ED &
        $\times$ &
        $\times$
        \\\midrule
		\TheName{} & 
        Expected accepted length & 
        On-policy & 
        Adaptive & 
        \checkmark{} &
        \checkmark{}
        \\
		\bottomrule
	\end{tabular}
    \vspace{-15px}
\end{table}

\section{Experiments}\label{sec:experiments}

\subsection{Experiment Setup}\label{sec:setup}

\paragraph{Baselines.}
We use two draft model architectures: \textbf{Eagle-3}~\citep{eagle3} and \textbf{DFlash}~\citep{dflash}.
For both architectures, we implement SFT to train them on target-generated sequences.
Based on DFlash architecture (our experimental results show that DFlash performs better than Eagle-3), we implement recent SOTA representative methods with public DFlash implementations as baselines: off-policy verification-aware  \textbf{D-PACE}~\citep{dpace}, \textbf{LK-Losses}~\citep{lk}, and on-policy \textbf{Draft-OPD}~\citep{draftopd}.
See Appendix~\ref{app:baselines} for implementation details of baselines.

\paragraph{Benchmarks.}
We include the following benchmarks:
(1) Mathematics: \textbf{GSM8K}~\citep{gsm8k}, \textbf{MATH-500}~\citep{math} and \textbf{AIME25}~\citep{aime25}; (2) Coding: \textbf{MBPP}~\citep{mbpp}, \textbf{HumanEval}~\citep{humaneval} and \textbf{SWE-Lite}~\citep{swebench}; (3) Multi-turn Chat: \textbf{MT-Bench}~\citep{mtbench}.
We report accepted length ($\tau$) and speedup ratio (SR) for benchmarks.
See Appendix~\ref{app:benchmarks} for details of benchmarks.

\paragraph{Implementations.} We use \textbf{Qwen3-4B / -8B}~\citep{qwen3} as target model (See Appendix~\ref{app:target} for generalization to other targets).
For SFT stage, we use the \textbf{Qwen3-4B-Instruct-100K} dataset released by DFlash~\citep{dflash} for draft model on Qwen3-4B.
We leverage Qwen3-8B to regenerate responses for prompts in Qwen3-4B-Instruct-100K, and use the generated sequences as SFT dataset for draft model on Qwen3-8B.
For on-policy training stage, we use 16K prompts released by Draft-OPD~\citep{draftopd} as training set.
While for off-policy baselines, we collect target-generated sequences on Qwen3-4B-Instruct-100K and the 16K prompts as training set.
We evaluate with temperature set to 0 and 1 respectively.
We set $L = 15$ for training and inference, and select $\beta=0.3$ for \TheName{} based on validation set.
We refer to previous works~\citep{dflash,draftopd} and disable the thinking mode of target models (see Appendix~\ref{app:thinking-mode} for results with thinking-mode-enabled).
See Appendix~\ref{app:implementation} for more implementation and hyperparameter selection.
\subsection{Overall Performance}

\begin{table}[t]
  \centering
  \caption{\label{tab:main-results}Decoding speedup ratio and average acceptance length ($\tau$) on Qwen3 models.}
  \vspace{-5pt}
  \small
  \setlength{\tabcolsep}{1.5pt}
  \begin{tabular}{ll*{7}{c}cc}
    \toprule
    \multicolumn{1}{c}{\multirow{3}{*}{\raisebox{-1.0ex}{\textbf{Model}}}}
    & \multicolumn{1}{c}{\multirow{3}{*}{\raisebox{-1.0ex}{\textbf{Method}}}}
    & \multicolumn{7}{c}{\textbf{Speedup Ratio (SR)}}
    & \multicolumn{2}{c}{\textbf{Mean}} \\
    \cmidrule(lr){3-9}
    \cmidrule(lr){10-11}
    & & \multicolumn{3}{c}{\textbf{Math}}
    & \multicolumn{3}{c}{\textbf{Code}}
    & \multicolumn{1}{c}{\textbf{Chat}}
    & \multirow{2}{*}{\raisebox{-0.45ex}{\textbf{SR}}}
    & \multirow{2}{*}{\raisebox{-0.45ex}{\(\boldsymbol{\tau}\)}} \\
    \cmidrule(lr){3-5}
    \cmidrule(lr){6-8}
    \cmidrule(lr){9-9}
    & & \textbf{GSM8K}
    & \textbf{MATH-500}
    & \textbf{AIME25}
    & \textbf{MBPP}
    & \textbf{HumanEval}
    & \textbf{SWE-Lite}
    & \textbf{MT-Bench}
    & & \\
    \midrule
    \rowcolor{orange!4}
    \multicolumn{11}{c}{Temperature $=0$} \\
    \midrule
    \multirow{6}{*}{Q3-4B} & EAGLE-3 & 1.99$\times$ & 1.83$\times$ & 1.79$\times$ & 1.78$\times$ & 1.84$\times$ & 1.62$\times$ & 1.74$\times$ & 1.80$\times$ & 5.32 \\
    & DFlash & 3.59$\times$ & 4.04$\times$ & 3.43$\times$ & 3.23$\times$ & 3.45$\times$ & 2.14$\times$ & 2.02$\times$ & 3.13$\times$ & 5.73 \\
	& D-PACE & 3.74$\times$ & 4.22$\times$ & 3.67$\times$ & 3.30$\times$ & 3.33$\times$ & 2.20$\times$ & 1.97$\times$ & 3.20$\times$ & 5.83 \\
	& LK-Losses & 3.92$\times$ & 4.48$\times$ & 3.88$\times$ & 3.35$\times$ & 3.52$\times$ & 2.22$\times$ & 2.00$\times$ & 3.34$\times$ & 6.01 \\
	& Draft-OPD & 4.24$\times$ & 4.74$\times$ & 4.00$\times$ & 3.42$\times$ & 3.52$\times$ & 2.16$\times$ & 2.06$\times$ & 3.45$\times$ & 6.23 \\
    \rowcolor[HTML]{F3F3F3}
    & \textbf{\TheName{}} & \textbf{4.62$\times$} & \textbf{5.13$\times$} & \textbf{4.43$\times$} & \textbf{3.72$\times$} & \textbf{3.96$\times$} & \textbf{2.43$\times$} & \textbf{2.25$\times$} & \textbf{3.79$\times$} & \textbf{6.94} \\
    \midrule
	% \addlinespace
    \multirow{6}{*}{Q3-8B} & EAGLE-3 & 1.94$\times$ & 1.81$\times$ & 1.79$\times$ & 1.69$\times$ & 1.89$\times$ & 1.73$\times$ & 1.63$\times$ & 1.78$\times$ & 5.20 \\
    & DFlash & 4.62$\times$ & 5.50$\times$ & 4.93$\times$ & 4.20$\times$ & 4.82$\times$ & 2.69$\times$ & 2.52$\times$ & 4.18$\times$ & 5.67 \\
	& D-PACE & 4.87$\times$ & 5.76$\times$ & 5.12$\times$ & 4.33$\times$ & 4.80$\times$ & 2.72$\times$ & 2.43$\times$ & 4.29$\times$ & 5.72 \\
	& LK-Losses & 5.14$\times$ & 6.06$\times$ & 5.38$\times$ & 4.52$\times$ & 4.96$\times$ & 2.88$\times$ & 2.69$\times$ & 4.52$\times$ & 6.03 \\
	& Draft-OPD & 5.40$\times$ & 6.36$\times$ & 5.62$\times$ & 4.78$\times$ & 5.17$\times$ & 3.06$\times$ & 2.87$\times$ & 4.75$\times$ & 6.34 \\
    \rowcolor[HTML]{F3F3F3}
    & \textbf{\TheName{}} & \textbf{5.82$\times$} & \textbf{6.82$\times$} & \textbf{6.01$\times$} & \textbf{5.14$\times$} & \textbf{5.52$\times$} & \textbf{3.23$\times$} & \textbf{3.08$\times$} & \textbf{5.09$\times$} & \textbf{7.08} \\
    \midrule
	\rowcolor{orange!4}
    \multicolumn{11}{c}{Temperature $=1.0$} \\
    \midrule
    \multirow{6}{*}{Q3-4B} & EAGLE-3 & 1.89$\times$ & 1.75$\times$ & 1.64$\times$ & 1.69$\times$ & 1.74$\times$ & 1.48$\times$ & 1.70$\times$ & 1.70$\times$ & 5.01 \\
    & DFlash & 3.31$\times$ & 3.24$\times$ & 2.44$\times$ & 3.03$\times$ & 3.25$\times$ & 1.90$\times$ & 1.90$\times$ & 2.72$\times$ & 5.34 \\
	& D-PACE & 3.49$\times$ & 3.37$\times$ & 2.52$\times$ & 2.97$\times$ & 3.13$\times$ & 1.93$\times$ & 1.82$\times$ & 2.75$\times$ & 5.42 \\
	& LK-Losses & 3.60$\times$ & 3.52$\times$ & 2.60$\times$ & 3.08$\times$ & 3.23$\times$ & 1.95$\times$ & 1.97$\times$ & 2.85$\times$ & 5.58 \\
	& Draft-OPD & 3.81$\times$ & 3.82$\times$ & 2.88$\times$ & 3.16$\times$ & 3.27$\times$ & 1.91$\times$ & 1.94$\times$ & 2.97$\times$ & 5.70 \\
    \rowcolor[HTML]{F3F3F3}
    & \textbf{\TheName{}} & \textbf{4.14$\times$} & \textbf{4.21$\times$} & \textbf{3.19$\times$} & \textbf{3.43$\times$} & \textbf{3.65$\times$} & \textbf{2.11$\times$} & \textbf{2.17$\times$} & \textbf{3.27$\times$} & \textbf{6.01} \\
    \midrule
	% \addlinespace
    \multirow{6}{*}{Q3-8B} & EAGLE-3 & 1.87$\times$ & 1.72$\times$ & 1.63$\times$ & 1.64$\times$ & 1.75$\times$ & 1.62$\times$ & 1.58$\times$ & 1.69$\times$ & 4.72 \\
    & DFlash & 4.15$\times$ & 4.23$\times$ & 3.09$\times$ & 3.57$\times$ & 4.01$\times$ & 2.08$\times$ & 2.16$\times$ & 3.33$\times$ & 4.53 \\
	& D-PACE & 4.27$\times$ & 4.39$\times$ & 3.20$\times$ & 3.69$\times$ & 3.94$\times$ & 2.17$\times$ & 2.10$\times$ & 3.39$\times$ & 4.64 \\
	& LK-Losses & 4.52$\times$ & 4.64$\times$ & 3.55$\times$ & 3.84$\times$ & 4.33$\times$ & 2.30$\times$ & 2.28$\times$ & 3.64$\times$ & 4.92 \\
	& Draft-OPD & 4.94$\times$ & 5.01$\times$ & 3.76$\times$ & 4.02$\times$ & 4.57$\times$ & 2.42$\times$ & 2.49$\times$ & 3.89$\times$ & 5.23 \\
    \rowcolor[HTML]{F3F3F3}
    & \textbf{\TheName{}} & \textbf{5.38$\times$} & \textbf{5.50$\times$} & \textbf{4.11$\times$} & \textbf{4.36$\times$} & \textbf{4.85$\times$} & \textbf{2.79$\times$} & \textbf{2.74$\times$} & \textbf{4.25$\times$} & \textbf{5.71} \\
    \bottomrule
  \end{tabular}
   \vspace{-15pt}
\end{table}

\paragraph{Main Results.}
According to Table~\ref{tab:main-results}, \TheName{} achieves the highest average SR and $\tau$ among all methods based on DFlash architecture, demonstrating the effectiveness of \TheName{}.
Specifically,
\textbf{(1)}
Compared with Eagle-3, DFlash has higher SR, because it generates multiple tokens of the draft in parallel, while Eagle-3 generates the draft tokens auto-regressively.
\textbf{(2)}
Compared with off-policy methods (DFlash, D-PACE and LK-Losses), on-policy methods (Draft-OPD and \TheName{}) obtain draft model with higher average SR and $\tau$, showing the effectiveness of learning from drafts generated by the draft model itself.
\textbf{(3)}
\TheName{} outperforms Draft-OPD, suggesting effectiveness of the proposed expected accepted length optimization method beyond on-policy training paradigm.
Besides Table~\ref{tab:main-results}, we also evaluate \TheName{} in the SGLang inference framework to assess deployment efficiency.
See Appendix~\ref{app:sglang} for the results.

\subsection{Ablation Study}\label{sec:ablation}
To evaluate the effectiveness of two techniques in Section~\ref{sec:optimization} (i.e. token-level acceptance rate expectation marginalization and over-sampled token penalty) and two properties of the gradient in Equation~\ref{eq:gradient2} (i.e. prefix-aware credit assignment and TVD), we design the following variants of \TheName{}:
(1) \textbf{w/o marg}: removes the token-level acceptance rate expectation marginalization and uses the gradient in Equation~\ref{eq:app-pf-gradient};
(2) \textbf{w/o penalty}: removes the over-sampled token penalty and uses the gradient in Equation~\ref{eq:gradient1};
(3) \textbf{\TheName-ED}: replaces the prefix-aware credit assignment with ED;
(4) \textbf{\TheName-KL}: replaces the TVD in gradient with forward / reverse KL for accepted and rejected tokens respectively, referring to Draft-OPD.
Since KL does not satisfy Equation~\ref{eq:sampled_tv}, we do not implement over-sampled token penalty in \TheName-KL.
So the only difference between \TheName-KL and w/o penalty is the TVD and KL.
See Appendix~\ref{app:variants} for implementation details.
We evaluate them on AIME25, HumanEval and MT-Bench, with DFlash architecture and Qwen3-4B as target model.

\begin{wraptable}{r}{0.57\textwidth}
		\centering
		\small
		\vspace{-10pt}
		\caption{\label{tab:ablation}Ablation study on Qwen3-4B, temperature = 0.} 
		\vspace{-5pt}
		\setlength{\tabcolsep}{4pt}
			\begin{tabular}{l*{6}{c}}
				\toprule
				\multirow{2}{*}{\raisebox{-0.45ex}{\textbf{Method}}}
				& \multicolumn{2}{c}{\textbf{AIME25}}
				& \multicolumn{2}{c}{\textbf{HumanEval}}
				& \multicolumn{2}{c}{\textbf{MT-Bench}} \\
				\cmidrule(lr){2-3}
				\cmidrule(lr){4-5}
				\cmidrule(lr){6-7}
				& \textbf{SR} & \(\boldsymbol{\tau}\)
				& \textbf{SR} & \(\boldsymbol{\tau}\)
				& \textbf{SR} & \(\boldsymbol{\tau}\) \\
				\midrule
				w/o marg
				& 3.12$\times$ & 5.72 & 3.20$\times$ & 5.78 & 1.74$\times$ & 5.09 \\
				w/o penalty
				& 4.32$\times$ & 7.70 & 3.80$\times$ & 6.90 & 2.12$\times$ & 5.32 \\
				\midrule
				\TheName-ED
				& 3.96$\times$ & 7.20 & 3.47$\times$ & 6.30 & 2.10$\times$ & 5.26 \\
				\TheName-KL
				& 4.19$\times$ & 7.56 & 3.66$\times$ & 6.67 & 2.12$\times$ & 5.33 \\
				\midrule
				\rowcolor[HTML]{F3F3F3}
				\textbf{\TheName{}}
				& \textbf{4.43$\times$} & \textbf{7.96}
				& \textbf{3.96$\times$} & \textbf{7.18}
				& \textbf{2.25$\times$} & \textbf{5.45} \\
				\bottomrule
			\end{tabular}
		\vspace{-10pt}
\end{wraptable}
We report the results in Table~\ref{tab:ablation}, and summarize observations as follows:
\textbf{(1)} Gradient estimator in Equation~\ref{eq:app-pf-gradient} underperforms naive DFlash SFT, indicating that it is not effective for on-policy training.
\textbf{(2)} Token-level expected acceptance rate marginalization provides the dominant enhancement.
w/o penalty substantially outperforms w/o marg, and performs similarly to \TheName{}, indicating that the marginalization is the dominant factor for enhancing the  draft model performance.
\textbf{(3)} Penalty for over-sampled token can provide additional enhancement.
The full \TheName{} still outperforms w/o penalty, showing that the penalty for over-sampled token can provide additional enhancement for draft model training.
\textbf{(4)} \TheName-ED slightly underperforms Draft-OPD, which also applies ED to weight the gradient, indicating that only TVD without prefix-aware credit assignment cannot enhance performance stably.
\textbf{(5)} \TheName-KL surpasses Draft-OPD, showing that prefix-aware credit assignment can effectively enhance performance even coupled with KL.
But \TheName-KL still underperforms \TheName{} and w/o penalty, indicating the effectiveness of whole \TheName{}, where prefix-aware credit assignment is coupled with TVD.

\subsection{Hyperparameter Sensitivity}\label{sec:hyperparameter}
\begin{wrapfigure}{tr}{0.39\textwidth}
	\centering
	\vspace{-15pt}
	\begin{subfigure}{0.39\textwidth}
        \includegraphics[width=\linewidth]{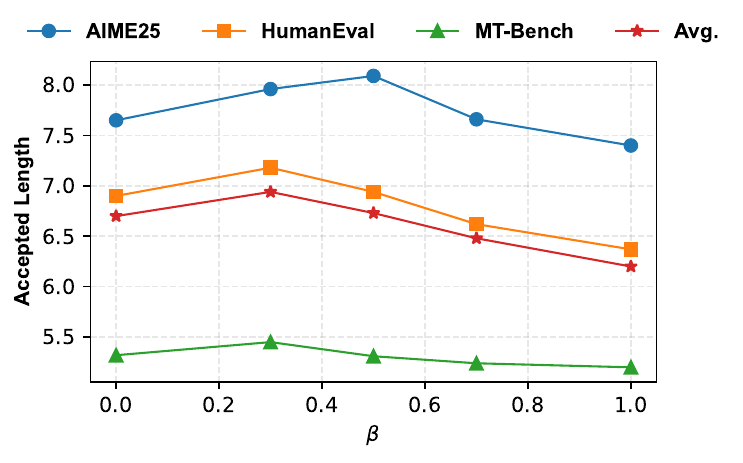}
	\end{subfigure}
	\vspace{-18pt}
	\caption{\label{fig:beta}Influence of $\beta$.}
	\vspace{-15pt}
\end{wrapfigure}
\textbf{Penalty Strength $\beta$.} We fix the draft length $L$ to 15 and set $\beta$ to $\{0.0, 0.3, 0.5, 0.7, 1.0\}$ respectively.
We report $\tau$ on AIME25, HumanEval, MT-Bench and the average metric across all seven benchmarks introduced in Section~\ref{sec:setup} in Figure~\ref{fig:beta}.
The results show that with $\beta$ from 0 to 1, the performance exhibits a trend of first increasing and then decreasing.
The optimal $\beta$ is different for different benchmarks, and $\beta=0.3$ achieves the best average performance.
See Section~\ref{sec:in-depth} about training dynamics for more analysis on influence of $\beta$ to training efficiency.

\begin{wrapfigure}{tr}{0.39\textwidth}
	\centering
	\vspace{-15pt}
	\begin{subfigure}{0.39\textwidth}
        \includegraphics[width=\linewidth]{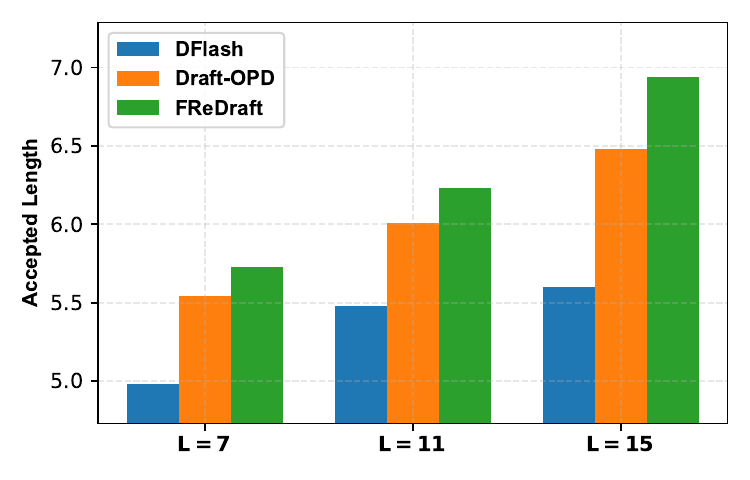}
	\end{subfigure}
	\vspace{-18pt}
	\caption{\label{fig:draft_length}Influence of $L$.}
	\vspace{-15pt}
\end{wrapfigure}

\textbf{Draft Length $L$.}
We fix $\beta$ to 0.3 and set $L$ to $\{7, 11, 15\}$ respectively during training and inference.
We report the average $\tau$ of DFlash, Draft-OPD and \TheName{} on all seven benchmarks in Figure~\ref{fig:draft_length}.
The results show that under larger $L$, \TheName{} demonstrates stronger advantage over baselines.
This is because larger $L$ introduces more gradient noise from later tokens in the draft, which are usually rejected or discarded in verification.
However, the prefix-aware credit assignment in \TheName{} adaptively assigns low weight to these tokens, leading to better advantage.

\subsection{In-Depth Analysis}\label{sec:in-depth}

\begin{figure}[htbp]
    \centering
    \includegraphics[width=\linewidth]{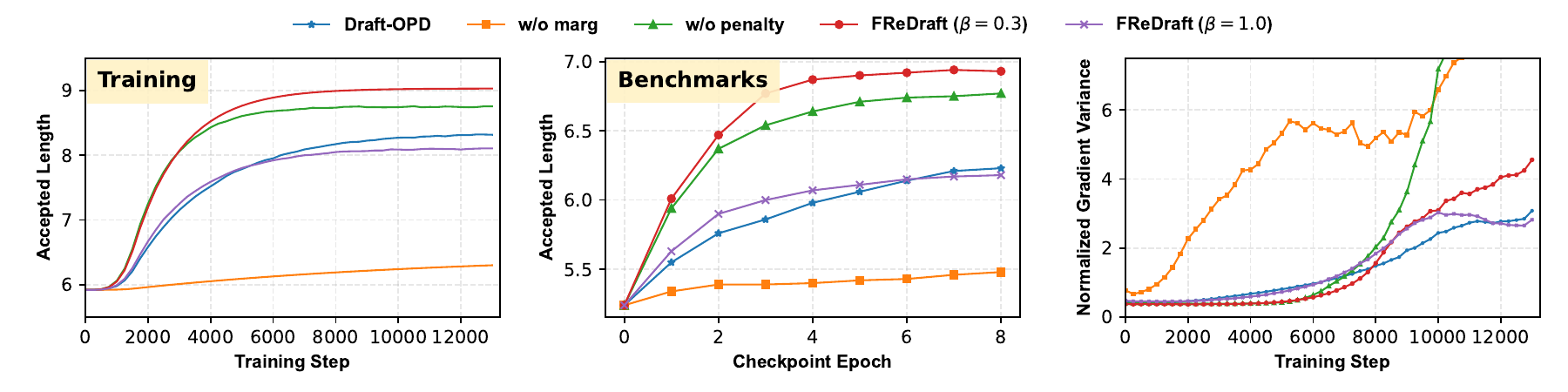}
	\vspace{-15pt}
    \caption{\label{fig:al-step}
    Average accepted length on training set, benchmarks and normalized gradient variance.
    }
    \vspace{-15pt}
\end{figure}
\paragraph{Training Dynamics Analysis.}
We report the training dynamics during on-policy stage to show the convergence efficiency of \TheName{}.
Specifically, we plot the average accepted length of the training-time collected draft blocks on training set and checkpoint at each epoch on evaluation benchmarks.
We also plot training-time normalized gradient variance (see Appendix~\ref{app:snr} for implementation details) during training.
The results are shown in Figure~\ref{fig:al-step}.
We summarize observations as follows:
\textbf{(1)} The accepted length of w/o marg improves extremely slowly.
We observe that 87.7\% of the tokens in the sampled draft receive zero gradient in w/o marg (i.e. gradient in Equation~\ref{eq:app-pf-gradient}).
Its gradient variance is large, leading to low training efficiency.
\textbf{(2)} Penalty for over-sampled tokens can further enhance convergence efficiency.
Compared with \TheName{}, though w/o penalty has similar accepted length curve during early training, its gradient variance increases much faster after middle training stage, and converges at lower accepted length, showing that penalty for over-sampled tokens can reduce the gradient noise from unsampled tokens and enhance convergence efficiency.
\textbf{(3)} \TheName{} and w/o penalty demonstrate higher convergence efficiency and final performance, showing the advantage of optimizing the draft expected accepted length.
\TheName{}, w/o penalty and Draft-OPD all leverage the target distribution to supervise draft model training.
The difference is that the gradient of \TheName{} and w/o penalty is derived from the expected accepted length of draft, while Draft-OPD is heuristically designed forward and reverse KL.
Therefore, this comparison isolates the effectiveness of expected accepted length maximization and shows that it can enhance final performance.
Moreover, compared with Draft-OPD, the gradient variance of \TheName{} and w/o penalty increases more slowly in early training stage.
See Appendix~\ref{app:wall-clock} for wall-clock time comparison of on-policy training.

\paragraph{Prefix Survival \& Token Acceptance Rate Distribution.}
\begin{wrapfigure}{r}{0.62\textwidth}
	\centering
	\vspace{-10pt}
	\begin{subfigure}{0.62\textwidth}
        \includegraphics[width=\linewidth]{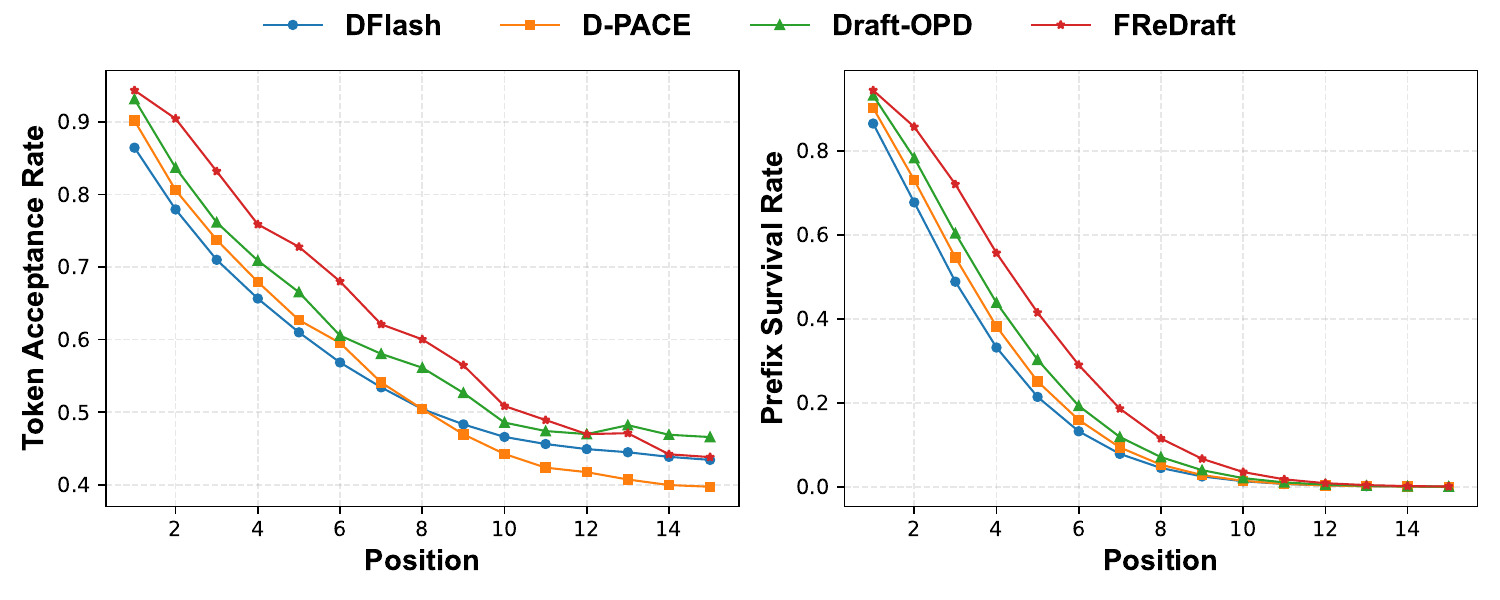}
	\end{subfigure}
	\vspace{-15pt}
	\caption{\label{fig:distribution}Token-acceptance \& prefix-survival rate distribution.}
	\vspace{-15pt}
\end{wrapfigure}
To reveal the internal mechanism by which \TheName{} improves draft accepted length during inference, we plot the average prefix survival rate and token-level acceptance rate distribution on different positions in all drafts collected on seven benchmarks.
According to the results on 15 draft-generated tokens in Figure~\ref{fig:distribution}, the prefix survival rate of all methods decreases as the position steps from 1 to 15.
However, the decreasing rate of \TheName{} is much slower than other methods, especially at positions 3--8.
The slowly-decreasing prefix survival rate of \TheName{} enables verification to proceed to later tokens in the draft, thus achieving larger accepted length and higher speedup ratio.
In terms of token-level acceptance rate, all methods exhibit extremely high acceptance rate for first several tokens, while \TheName{} still shows a slight advantage. 
For positions 3--8, \TheName{} demonstrates significant acceptance-rate advantage over other methods, which is consistent with the observation of prefix survival rate.
For later positions, some methods show higher token acceptance rate than \TheName{}.
But due to first-rejection truncation, verification process rarely reaches these positions, leading to that improvements in acceptance at these positions have limited impact on accepted length.

\section{Conclusion}\label{sec:conclusion}
We propose \TheName{}, an on-policy draft-model training method that directly maximizes draft expected accepted length to model the first-rejection truncation property of verification.
For efficient optimization, we introduce token-level acceptance-rate marginalization and over-sampled token penalty, enabling unbiased low-variance gradient estimation.
Experiments across benchmarks and targets show that
\TheName{} consistently improves accepted length and speedup over strong baselines.

\clearpage
\bibliography{iclr2027_conference}
\bibliographystyle{iclr2027_conference}

\clearpage
\appendix
\section{Detailed Discussion about Speculative Draft Models}\label{app:related-works}

\subsection{Draft Model Architectures}

\paragraph{Independent autoregressive drafters.}
The original speculative decoding and speculative sampling algorithms use a small, independently parameterized language model to generate a draft autoregressively, after which the target model verifies the complete draft in parallel~\citep{sd1,sd2}.
The draft and target can be developed and deployed independently, and the same verification rule also supports tree-structured proposals~\citep{specinfer}.
The main limitation is that autoregressive draft generation places a serial model invocation on the critical path for every proposed token.
Consequently, increasing the draft length improves the amount of work amortized by one verification call, but also increases drafting latency.

\paragraph{Self-drafting and feature-based drafters.}
Another line of work reuses computation or representations from the target model.
Draft\&Verify skips selected target layers to construct a faster self-draft~\citep{zhang2024draft}, while QSpec uses a complementary low-precision execution of the target as the drafter~\citep{qspec}.
Medusa attaches multiple prediction heads to a frozen target model and produces a tree of candidates for several future positions~\citep{medusa}.
The EAGLE family instead predicts future target-model features and then maps those features to token distributions~\citep{eagle,eagle2}.
Eagle-3~\citep{eagle3} improves candidate-tree construction and training-time feature alignment.
These methods reduce or share draft computation, but their draft distributions still need to be aligned with the target distribution for high acceptance.

\paragraph{Parallel block drafters.}
DFlash formulates drafting as block diffusion: conditioned on features from the target-model prefix, a lightweight draft network predicts an entire masked block in parallel~\citep{dflash}.
DSpark likewise uses semi-autoregressive generation with a confidence-aware schedule to expose more parallelism during draft construction~\citep{dspark}.
Parallel block drafters remove most of the serial drafting cost and therefore make accepted length especially important: once a block has been produced, accepting a longer prefix adds little drafting overhead.
Our experiments use DFlash because it provides this efficient parallel drafting path.  The objective proposed in this paper is nevertheless defined over the induced draft distribution and is not tied to a particular neural architecture.

\subsection{Draft Model Training Methods}\label{app:training-method}

\paragraph{Supervised fine-tuning.}
A common recipe trains the draft model with token-level cross-entropy on sequences generated by the target model~\citep{eagle3,dflash,specforge}.
This supervised finetuning (SFT) method is stable.
However, it observes target prefixes during training whereas deployment conditions the draft model on its own proposals, leading to exposure bias.

\paragraph{Distribution distillation.}
Compared with SFT, white-box distillation provides the target distribution over vocabulary as supervision to train the draft, thereby conveying more information than a single hard token and typically achieves better draft model performance~\citep{gui2024boosting,ouyang2024temperature}.
For example, DistillSpec studies draft-model knowledge distillation with different data sources and distribution-divergence objectives~\citep{distillspec}.
OSD~\citep{online}, OnlineSpec~\citep{qian2026when} and ReSpec~\citep{respec} further model an online training problem for continuous adaptation of the draft model during deployment.
They leverage the vocabulary distribution output by target model during deployment to train the draft model continuously.

\paragraph{Verification-aware training.}
Recent methods begin to explore incorporating property of speculative verification process into draft model training, rather than merely letting the draft model learn to imitate target-generated tokens or distributions.
For example, LK-Losses proposes to directly optimize token-level acceptance rate during verification and derives the objective as minimizing the TVD between draft model and target model distribution~\citep{lk}.
However, LK-Losses optimizes the draft distribution at each position with static exponential decay as weight, without modeling the influence from earlier tokens to later tokens.
Draft accepted length is a metric directly related to the speedup ratio of speculative decoding given other draft model, target model and device configurations~\citep{qspec,zhang2024draft}.
Therefore, some recent methods take accepted length into account when designing their training method~\citep{bhansali2025draft,vat,specdec,bachmann2025judge,garipov2026autojudge}.
PARD-2~\citep{pard2} observes that cumulative confidence of target model over the preceding prefix is closely correlated to the token-level acceptance rate and it weights the gradient of tokens at different positions with the cumulative confidence.
Similarly, D-PACE~\citep{dpace} observes that the draft-model summed cumulative confidence on target-generated sequence chunks is empirically related to the average accepted length of drafts, and D-PACE derives position weight in training data according to this surrogate.
VSD~\citep{vsd} proposes a variational objective as lower bound of draft tree utility and implement Expectation-Maximization algorithm to optimize the objective. 
Besides these, two concurrent works VAT~\citep{vat} and WTV~\citep{wtv} also take verification information into training objective.
Despite effectiveness of above verification-aware training method compared with naive SFT, these methods, however, optimize a metric of sequences under target-distribution as surrogate, which does not match ``first-rejection truncation'' in practical verification.
This results in exposure bias and gap between optimized objective and real verification process.
Take VSD as an example, although it uses the draft model to propose training sequences, it further leverages the target model to filter and correct training samples for subsequent training process.
Therefore, the trial-and-error samples, which are important for mitigating exposure bias, are excluded in training.
Moreover, the optimized surrogate may not necessarily lead to later ``first-rejection".
For example, D-PACE optimizes summed cumulative draft confidence, which makes draft model diverge from target when target model itself is uncertain.
In comparison, \TheName{} directly optimizes the expected accepted length of draft generated by draft model itself, which naturally matches on-policy training paradigm, thus achieving better performance.

\paragraph{On-policy paradigm.}
On-policy paradigm trains the draft model on its own generated sequences, matching practical verification process and eliminating exposure bias.
An early work is GTO~\citep{gto} designed specifically for tree-structured draft (e.g. Eagle-3).
It adopts semi-on-policy training by combining reinforcement learning (RL) with supervised finetuning.
However, GTO implements its RL training recipe referring to GRPO~\citep{deepseekmath}, which leads to coarse-grained credit assignment.
On-policy distillation (OPD) combines the advantage of on-policy training and dense supervision of distillation, thus becoming a widely-adopted training paradigm in LLM post-training stage~\citep{v4,mimov2,k3,glm5}.
Inspired by the success of OPD in LLM post-training, Draft-OPD proposes to train the draft model on drafts collected by itself during speculative decoding~\citep{draftopd}.
Its released training recipe applies a forward-KL-style loss to accepted/committed tokens and a sampled reverse-KL estimator to invalidated draft suffixes, together with a static exponential decay over draft positions.
On-policy collection removes the train--inference prefix mismatch, but the token-level divergences and static position decay do not represent the realized survival probability of each prefix, leading to inefficient training signal allocation~\citep{dflash,dpace}.
In contrast, \TheName{} retains on-policy collection while deriving its credit assignment from the expected accepted length itself.
The resulting weight of a proposal depends on both the acceptance of its prefix and the future accepted-length contributions that the proposal enables.

\section{Implementation Details}\label{app:implementation}

\subsection{Block-Level Estimator Construction}\label{app:block}
Algorithm~\ref{alg:block-estimator} expands the estimator-computation step in Algorithm~\ref{alg:training}.  The implementation uses a minimization surrogate whose negative gradient equals the ascent estimator in Equation~\ref{eq:gradient2}.
Here $\operatorname{sg}(\cdot)$ denotes stop-gradient.
The reverse cumulative sum used to compute $w_t$ excludes the current position, so $\sum_{k=t+1}^{L}A_kS_{k-1}=0$ for $t = L$; this is the implementation of the ``beyond position $t$'' term in Proposition~\ref{prop:marginalization}.

\begin{algorithm}[ht]
    \small
    \caption{Block-Level Construction of $\widehat{\mathbf{g}}_{\beta}$}
    \label{alg:block-estimator}
    \begin{algorithmic}[1]
        \REQUIRE Complete collected draft blocks $\mathcal{B}$, frozen target
        model $p$, draft model $q_{\theta}$, penalty coefficient $\beta$
        \ENSURE Batch estimator $\widehat{\mathbf{g}}_{\beta}$

        \STATE Initialize $\mathcal{L}\leftarrow0$ and $M\leftarrow0$
        \FOR{each $\mathbf{y}=(y_1,\ldots,y_L)\in\mathcal{B}$}
            \FOR{$t=1,\ldots,L$}
                \STATE Obtain $p_t$, $q_{\theta,t}$, and the sampled token $y_t$
                from verification / replay
                \STATE $\alpha_t\leftarrow
                \exp\!\left(\min\{0,\log p_t(y_t)-\log q_{\theta,t}(y_t)\}\right)$
                \STATE $A_t\leftarrow1-D^{(K)}_{\mathrm{TV}}(p_t,q_{\theta,t})$
            \ENDFOR
            \STATE $S_0\leftarrow1$ and
            $S_{t-1}\leftarrow\prod_{j=1}^{t-1}\alpha_j$ for $t=2,\ldots,L$
            \FOR{$t=1,\ldots,L$}
                \STATE $w_t\leftarrow
                (\sum_{k=t+1}^{L}A_kS_{k-1})\mathbbm{1}_{p_t(y_t)>q_{\theta,t}(y_t)}
                -(\beta S_{t-1})\mathbbm{1}_{p_t(y_t)<q_{\theta,t}(y_t)}$
                \STATE $\ell_t\leftarrow
                -\operatorname{sg}(w_t)\log q_{\theta,t}(y_t)
                +\operatorname{sg}((1-\beta)S_{t-1})D^{(K)}_{\mathrm{TV}}(p_t,q_{\theta,t})$
            \ENDFOR
            \STATE $\mathcal{L}\leftarrow\mathcal{L}+\sum_{t=1}^{L}\ell_t$;
            $M\leftarrow M+1$
        \ENDFOR
        \STATE \textbf{return}
        $\widehat{\mathbf{g}}_{\beta}=-\nabla_{\theta}(\mathcal{L}/M)$
    \end{algorithmic}
\end{algorithm}

\subsection{Memory-efficient distribution representation}\label{app:topk}
On-policy training requires target-distribution information for collected draft proposals, including rejected tokens and their invalidated suffixes.
For $N$ collected proposal positions and vocabulary $\mathcal{V}$, retaining the full target distributions requires $\mathcal{O}(N|\mathcal{V}|)$ memory.
This storage cost can become prohibitive and cause out-of-memory errors when collecting and replaying many draft blocks.
Following the use of target Top-K entries in on-policy distillation~\citep{gkd,fu2026revisiting}, we compress the stored target distributions to reduce this cost to $\mathcal{O}(NK)$.
Specifically, the implementation is as follows:

\paragraph{Implementation.}
At position $t$, we retain the target top-$K$ tokens and
the sampled draft token:
\begin{equation*}
	\mathcal{K}_t
	:=
	\operatorname{TopK}(p_t)\cup\{y_t\},
	\qquad
	\mathcal{T}_t
	:=
	\mathcal{V}\setminus\mathcal{K}_t.
\end{equation*}
During replay, we gather the draft probabilities at the same tokens, using the original vocabulary-wide normalization rather than renormalizing over $\mathcal{K}_t$.
All omitted tokens are aggregated into one residual category with target and draft masses
\begin{equation}
	\begin{aligned}
		\varepsilon_{p,t}
		&:=
		1-\sum_{v\in\mathcal{K}_t}p_t(v),\\
		\varepsilon_{q,t}
		&:=
		1-\sum_{v\in\mathcal{K}_t}q_{\theta,t}(v),\\
		\varepsilon_t
		&:=
		\min\{\varepsilon_{p,t},\varepsilon_{q,t}\}.
	\end{aligned}
	\label{eq:topk-tail-masses}
\end{equation}
The resulting TVD on the common retained-token-plus-tail partition is
\begin{equation*}
	\widetilde{D}^{(K)}_{\mathrm{TV}}(p_t,q_{\theta,t})
	:=
	\frac{1}{2}
	\left[
	\sum_{v\in\mathcal{K}_t}
	\left|p_t(v)-q_{\theta,t}(v)\right|
	+
	\left|\varepsilon_{p,t}-\varepsilon_{q,t}\right|
	\right].
\end{equation*}
This representation preserves normalization and all retained
probability mass.
The draft tail mass $\varepsilon_{q,t}$ remains differentiable;
neither it nor the retained draft probabilities are detached.
Because $y_t\in\mathcal{K}_t$, the sampled-token acceptance ratio and
under-/over-sampling indicators are computed from exact probabilities.
The compression introduces no additional target-model forward pass.

For brevity, write
\[
D_t:=D_{\mathrm{TV}}(p_t,q_{\theta,t}),
\qquad
\widetilde{D}_t
:=\widetilde{D}^{(K)}_{\mathrm{TV}}(p_t,q_{\theta,t}),
\]
and
\[
A_t:=1-D_t,
\qquad
\widetilde{A}_t:=1-\widetilde{D}_t.
\]
Here $\widetilde{A}_t$ is the approximation used for the marginalized
acceptance coefficient in Algorithm~\ref{alg:block-estimator}.

Next, we provide theoretical analysis and empirical analysis to show that the approximation error is negligible in practice.

\paragraph{Theoretical bounds.}
The approximation error is controlled by the omitted probability mass, independently of vocabulary size.
Throughout the analysis, the target distribution, sampled prefix, and realized retained set are held fixed during differentiation.
The gradient statements apply away from the nondifferentiable equality boundaries of the absolute-value terms and concern tail aggregation in exact arithmetic.

\begin{proposition}[TVD and logit-gradient error under tail aggregation]
	\label{prop:topk-tvd-bounds}
	Let $q_{\theta,t}=\operatorname{softmax}(z_t)$ and define
	$\varepsilon_t$ as in Equation~\ref{eq:topk-tail-masses}.
	Then
	\begin{align}
		0
		\le
		D_t-\widetilde{D}_t
		=
		\widetilde{A}_t-A_t
		&\le \varepsilon_t,
		\label{eq:topk-value-bound}\\
		\left\|
		\nabla_{z_t}D_t-\nabla_{z_t}\widetilde{D}_t
		\right\|_1
		&\le 2\varepsilon_t,
		\label{eq:topk-gradient-l1}\\
		\left\|
		\nabla_{z_t}D_t-\nabla_{z_t}\widetilde{D}_t
		\right\|_2
		&\le \sqrt{2}\,\varepsilon_t.
		\label{eq:topk-gradient-l2}
	\end{align}
	Furthermore, with
	$\mathbf{J}_t:=\partial z_t/\partial\theta$,
	\begin{equation}
		\left\|
		\nabla_\theta D_t-\nabla_\theta\widetilde{D}_t
		\right\|_2
		\le
		\sqrt{2}\,
		\|\mathbf{J}_t\|_{\mathrm{op}}\,
		\varepsilon_t,
		\label{eq:topk-parameter-bound}
	\end{equation}
	where $\|\cdot\|_{\mathrm{op}}$ denotes the Euclidean operator norm.
\end{proposition}

Proposition~\ref{prop:topk-tvd-bounds} bounds both quantities affected by compression: the marginalized acceptance coefficient and the dense TVD gradient.
Importantly, the bounds depend on the \emph{smaller} omitted mass, rather than the sum of the target and draft tail masses.
They hold pointwise for every realized retained set, including when that set contains the sampled token.

We can also bound the perturbation of the complete block-level gradient estimator.
Let $\widehat{\mathbf{g}}_\beta$ denote the full-vocabulary estimator in Equation~\ref{eq:gradient2}, and let $\widehat{\mathbf{g}}_\beta^{(K)}$ denote its compressed implementation in Algorithm~\ref{alg:block-estimator}.

\begin{corollary}[Compression error of the block-level estimator]
	\label{cor:topk-estimator-bias}
	Suppose that, for all relevant positions, prefixes, and tokens,
	\begin{equation}
		\left\|
		\nabla_\theta\log q_{\theta,t}(v)
		\right\|_2
		\le C.
		\label{eq:topk-score-assumption}
	\end{equation}
	For any $\beta\in[0,1]$, evaluating the full and compressed estimators on the same sampled block gives
	\begin{equation}
		\left\|
		\widehat{\mathbf{g}}_\beta^{(K)}
		-
		\widehat{\mathbf{g}}_\beta
		\right\|_2
		\le
		C\sum_{k=1}^{L}
		(k-\beta)S_{k-1}\varepsilon_k.
		\label{eq:topk-pathwise-estimator-bound}
	\end{equation}
	Consequently, under on-policy draft sampling with the decoding-context distribution held fixed during differentiation,
	\begin{equation}
		\left\|
		\mathbb{E}
		\left[\widehat{\mathbf{g}}_\beta^{(K)}\right]
		-
		\nabla_\theta J(\theta;\bar{\theta})
		\right\|_2
		\le
		C\,\mathbb{E}
		\left[
		\sum_{k=1}^{L}
		(k-\beta)S_{k-1}\varepsilon_k
		\right],
		\label{eq:topk-estimator-bias-bound}
	\end{equation}
	where $J(\theta;\bar{\theta})$ is the fixed-context-distribution objective defined in Appendix~\ref{app:direct-gradient}.
\end{corollary}

Thus, the full-vocabulary estimator retains its exact expectation-preservation guarantee, while the compressed implementation has a tail-mass-dependent approximation error.
The proofs of both results are provided after the following empirical discussion.

\paragraph{Empirical evidence.}
We use $K=16$ for \TheName{} and its corresponding loss variants.
Empirically, the target top-$16$ entries cover $99.7\%$ of the target probability mass and $97.6\%$ of the draft probability mass on average.
Additionally retaining the sampled token can only decrease the omitted masses.
Let $\widehat{\mathbb{E}}$ denote averaging over the distributions used for these coverage measurements.
Applying Proposition~\ref{prop:topk-tvd-bounds} pointwise and averaging gives
\begin{align*}
	\widehat{\mathbb{E}}
	\left[D_t-\widetilde{D}_t\right]
	&\le
	\widehat{\mathbb{E}}[\varepsilon_t]
	\le
	\widehat{\mathbb{E}}[\varepsilon_{p,t}],
	\\
	\widehat{\mathbb{E}}
	\left[
	\left\|
	\nabla_{z_t}D_t
	-
	\nabla_{z_t}\widetilde{D}_t
	\right\|_2
	\right]
	&\le
	\sqrt{2}\,
	\widehat{\mathbb{E}}[\varepsilon_t]
	\le
	\sqrt{2}\,
	\widehat{\mathbb{E}}[\varepsilon_{p,t}].
\end{align*}
Using the reported rounded target-tail mass of $0.003$, these inequalities give approximate upper bounds of $3.0\times10^{-3}$ for the mean TVD error and $\sqrt{2}\times0.003\approx4.24\times10^{-3}$ for the mean Euclidean logit-gradient error.
Equivalently, the average marginalized acceptance rate is overestimated by at most approximately $0.3$ percentage points.
These measurements support using $K=16$ as a memory-efficient approximation with small average absolute error.

\paragraph{Proofs.}

\begin{proof}[Proof of Proposition~\ref{prop:topk-tvd-bounds}]
	Suppress the position index and write
	$q=\operatorname{softmax}(z)$.
	If the tail is empty, the full and compressed expressions coincide,
	so all bounds hold with zero error.
	Otherwise, define the positive and negative discrepancies within
	the tail:
	\[
	U:=\sum_{v\in\mathcal{T}}(q(v)-p(v))_+,
	\qquad
	V:=\sum_{v\in\mathcal{T}}(p(v)-q(v))_+,
	\]
	where $(a)_+:=\max\{a,0\}$.
	The retained terms are identical in $D$ and $\widetilde{D}$, while
	\[
	\sum_{v\in\mathcal{T}}|q(v)-p(v)|=U+V,
	\qquad
	\varepsilon_q-\varepsilon_p=U-V.
	\]
	Therefore,
	\begin{equation*}
		D-\widetilde{D}
		=
		\frac{1}{2}\left(U+V-|U-V|\right)
		=
		\min\{U,V\}
		\le
		\min\{\varepsilon_q,\varepsilon_p\}
		=
		\varepsilon.
	\end{equation*}
	Here $U\le\varepsilon_q$ and $V\le\varepsilon_p$.
	Since $\widetilde{A}-A=D-\widetilde{D}$, this proves
	Equation~\ref{eq:topk-value-bound}.
	
	For the gradient, define
	\[
	s_v:=\operatorname{sign}(q(v)-p(v)),
	\qquad
	\sigma:=\operatorname{sign}(\varepsilon_q-\varepsilon_p).
	\]
	Away from the equality boundaries, differentiating the full and
	compressed TVDs gives
	\begin{equation*}
		\Delta_z
		:=
		\nabla_zD-\nabla_z\widetilde{D}
		=
		\frac{1}{2}
		\sum_{v\in\mathcal{T}}
		(s_v-\sigma)\nabla_zq(v).
	\end{equation*}
	Only tail tokens whose individual discrepancy has the opposite sign
	to the aggregate tail discrepancy contribute to this difference.
	Let
	\[
	\mathcal{M}
	:=
	\left\{
	v\in\mathcal{T}:
	(q(v)-p(v))(\varepsilon_q-\varepsilon_p)<0
	\right\},
	\qquad
	m:=q(\mathcal{M}).
	\]
	Then
	\begin{equation}
		\Delta_z
		=
		-\sigma\sum_{v\in\mathcal{M}}\nabla_zq(v)
		=
		-\sigma\nabla_zq(\mathcal{M}).
		\label{eq:topk-subset-gradient}
	\end{equation}
	
	If $\varepsilon_q>\varepsilon_p$, every token in $\mathcal{M}$
	satisfies $q(v)<p(v)$, and hence
	\[
	m=q(\mathcal{M})
	\le p(\mathcal{M})
	\le\varepsilon_p.
	\]
	If $\varepsilon_q<\varepsilon_p$, then
	$m\le q(\mathcal{T})=\varepsilon_q$.
	Thus, in either case,
	\begin{equation*}
		m\le\min\{\varepsilon_p,\varepsilon_q\}
		=\varepsilon.
	\end{equation*}
	
	The softmax Jacobian gives
	\begin{equation*}
		\frac{\partial q(\mathcal{M})}{\partial z(v)}
		=
		q(v)\bigl(\mathbbm{1}_{v\in\mathcal{M}}-m\bigr)
		=
		\begin{cases}
			q(v)(1-m), & v\in\mathcal{M},\\
			-mq(v), & v\notin\mathcal{M}.
		\end{cases}
	\end{equation*}
	Using Equation~\ref{eq:topk-subset-gradient},
	\begin{align*}
		\|\Delta_z\|_1
		&=
		(1-m)\sum_{v\in\mathcal{M}}q(v)
		+
		m\sum_{v\notin\mathcal{M}}q(v)
		\\
		&=2m(1-m)
		\le 2\varepsilon.
	\end{align*}
	Similarly,
	\begin{align*}
		\|\Delta_z\|_2^2
		&=
		(1-m)^2\sum_{v\in\mathcal{M}}q(v)^2
		+
		m^2\sum_{v\notin\mathcal{M}}q(v)^2
		\\
		&\le
		(1-m)^2m^2+m^2(1-m)^2
		\\
		&=2m^2(1-m)^2
		\le 2\varepsilon^2.
	\end{align*}
	This proves Equations~\ref{eq:topk-gradient-l1}
	and~\ref{eq:topk-gradient-l2}.
	
	Finally, the chain rule yields
	\[
	\nabla_\theta D_t-\nabla_\theta\widetilde{D}_t
	=
	\mathbf{J}_t^\top
	\left(
	\nabla_{z_t}D_t-\nabla_{z_t}\widetilde{D}_t
	\right).
	\]
	Taking the Euclidean norm and applying
	Equation~\ref{eq:topk-gradient-l2} proves
	Equation~\ref{eq:topk-parameter-bound}.
\end{proof}

\begin{proof}[Proof of Corollary~\ref{cor:topk-estimator-bias}]
	Define
	\[
	\delta_t:=D_t-\widetilde{D}_t,
	\qquad
	\mathbf{e}_t
	:=
	\nabla_\theta D_t-\nabla_\theta\widetilde{D}_t.
	\]
	By Proposition~\ref{prop:topk-tvd-bounds}, $0\le\delta_t\le\varepsilon_t$.
	The parameter-space counterpart of Equation~\ref{eq:topk-subset-gradient}, together with the score bound in Equation~\ref{eq:topk-score-assumption}, gives
	\begin{align*}
		\|\mathbf{e}_t\|_2
		&=
		\left\|
		\sum_{v\in\mathcal{M}_t}
		q_{\theta,t}(v)
		\nabla_\theta\log q_{\theta,t}(v)
		\right\|_2
		\\
		&\le
		Cq_{\theta,t}(\mathcal{M}_t)
		\le C\varepsilon_t.
	\end{align*}
	
	Recall that $G_t=\mathbbm{1}_{p_t(y_t)>q_{\theta,t}(y_t)}
	\nabla_\theta\log q_{\theta,t}(y_t)$ and that $\widetilde{G}_t$ is the corresponding over-sampled-token score term.
	Expanding Equation~\ref{eq:gradient2}, the full-vocabulary estimator is
	\[
	\widehat{\mathbf{g}}_\beta
	=
	\sum_{t=1}^{L}
	\left[
	\left(
	\sum_{k=t+1}^{L}A_kS_{k-1}
	\right)G_t
	-
	(1-\beta)S_{t-1}\nabla_\theta D_t
	-
	\beta S_{t-1}\widetilde{G}_t
	\right].
	\]
	The compressed implementation replaces $A_k$ by $\widetilde{A}_k$ and $D_t$ by $\widetilde{D}_t$.
	The quantities $S_{t-1}$, $G_t$, and $\widetilde{G}_t$ remain unchanged because the sampled-token probabilities are exact.
	Since $\widetilde{A}_k-A_k=\delta_k$, subtracting the two estimators on the same sampled block gives
	\begin{equation*}
		\begin{aligned}
			\widehat{\mathbf{g}}_\beta^{(K)}
			-
			\widehat{\mathbf{g}}_\beta
			={}&
			(1-\beta)
			\sum_{t=1}^{L}S_{t-1}\mathbf{e}_t
			\\
			&+
			\sum_{t=1}^{L}\sum_{k=t+1}^{L}
			S_{k-1}\delta_kG_t.
		\end{aligned}
	\end{equation*}
	The score assumption implies $\|G_t\|_2\le C$.
	Therefore,
	\begin{align}
		\left\|
		\widehat{\mathbf{g}}_\beta^{(K)}
		-
		\widehat{\mathbf{g}}_\beta
		\right\|_2
		&\le
		C(1-\beta)\sum_{t=1}^{L}S_{t-1}\varepsilon_t
		+
		C\sum_{t=1}^{L}\sum_{k=t+1}^{L}
		S_{k-1}\varepsilon_k
		\nonumber\\
		&=
		C\sum_{k=1}^{L}(k-\beta)S_{k-1}\varepsilon_k.
		\label{eq:topk-estimator-bound-proof}
	\end{align}
	This proves Equation~\ref{eq:topk-pathwise-estimator-bound}.
	
	Under the stated on-policy and fixed-context-distribution assumptions, Corollary~\ref{cor:hybrid} gives
	\[
	\mathbb{E}[\widehat{\mathbf{g}}_\beta]
	=
	\nabla_\theta J(\theta;\bar{\theta}).
	\]
	Taking expectations in Equation~\ref{eq:topk-estimator-bound-proof} and using $\|\mathbb{E}[X]\|_2\le\mathbb{E}[\|X\|_2]$ proves Equation~\ref{eq:topk-estimator-bias-bound}.
\end{proof}

\subsection{Hyperparameters and $\beta$ Selection}\label{app:hyperparameter}
Table~\ref{tab:training-hyperparameters} summarizes the on-policy training configuration used for \TheName{} and its loss ablations. 
Unless a sensitivity experiment explicitly varies $L$ or $\beta$, the same values are used for all methods in the comparison. 
The target is frozen throughout this stage and only the DFlash draft-network parameters are optimized.
In Table~\ref{tab:training-hyperparameters}, $L=15$ is a widely used experiment setting in previous works, including DFlash and Draft-OPD (Notably, the two works set DFlash block size as 16, with one anchor token generated by target model. So the draft length is also 15).
To select penalty coefficient $\beta$, we randomly sample 10\% prompts from the 16K prompts released by Draft-OPD as validation set.
Then with the selected $\beta = 0.3$, we use all 16K prompts to conduct \TheName{} on-policy training to obtain results in Section~\ref{sec:experiments}.

\begin{table}[ht]
    \centering
    \small
    \caption{On-policy training hyperparameters.}
    \label{tab:training-hyperparameters}
    \setlength{\tabcolsep}{6pt}
    \begin{tabular}{ll}
        \toprule
        \textbf{Hyperparameter} & \textbf{Value} \\
        \midrule
        Target models & Qwen3-4B and Qwen3-8B \\
        Draft architecture & 5-layer DFlash \\
        Draft length $L$ & 15 \\
        Target distribution retained support $K$ & 16 \\
        Penalty coefficient $\beta$ & 0.3 \\
        ED coefficient (\TheName-ED only) & 0.8 \\
        Training epochs & 8 \\
        Prompt batch size & 12 \\
        Micro-batch size per draft worker & 1 \\
        Update-accumulation steps & 10 \\
        Learning rate & $2\times10^{-4}$ \\
        Training proposal sampling & temperature 1.0, top-$p=1.0$ \\
        \bottomrule
    \end{tabular}
\end{table}

\subsection{Other Key Implementation Details}\label{app:other-implementation}

\paragraph{Draft initialization and DFlash replay.}
We initialize from the SFT DFlash checkpoints described in Section~\ref{sec:setup}.
For both Qwen3 targets, the DFlash drafter contains five transformer layers and consumes hidden states from target layers $\{1,9,17,25,33\}$.
Its configured block size is 16: the first entry is an already available anchor and the remaining 15 entries are future-token proposals predicted in parallel. 
For the first speculative block, response token zero---sampled during target prefill---is the anchor; subsequent blocks are anchored at the preceding verification boundary.
This reproduces the layout used by DFlash inference.

During rollout, we record every verification boundary, every proposed token,
its offset within the block, its target log-probability, and the retained
target-distribution entries.  Accepted proposals are present in the committed
response, whereas the first rejected proposal and its remaining invalidated
suffix are carried in rejected-suffix metadata.  During training replay, the
two streams are joined by
$(\text{sample},\text{anchor},\text{offset})$ to reconstruct the original
on-policy block.  We use response-anchor stride one, disable random anchor
subsampling, and do not cap the number of rejected-suffix tokens.  A block is
used only if all proposal offsets are recovered; this prevents a truncated
suffix from being interpreted as a shorter draft.

\paragraph{Distribution capture and numerical stability.}
The frozen target logits used for the loss are captured from the verification/replay computation already required to obtain DFlash target features. 
The draft probabilities at the same token IDs are gathered from the trainable draft logits, and the remaining draft mass is computed by subtraction.
Consequently, constructing the \TheName{} loss requires no additional target-model forward pass relative to the on-policy DFlash replay.
Probability ratios are evaluated in log space as $\exp(\min\{0,\log p_t(y_t)-\log q_{\theta,t}(y_t)\})$. 
The frozen target
and the draft network use the same tokenizer and vocabulary.

\paragraph{Training and evaluation systems.}
Training is implemented on top of the VERL on-policy distillation pipeline with FSDP for the draft learner and SGLang for DFlash rollout.
In the Qwen3-4B configuration, three A100-80GB GPUs host draft-model workers and one A100-80GB GPU hosts the frozen target worker.
The target model is always run in evaluation mode, and thinking mode is disabled unless explicitly stated.
For evaluation, all methods use the same tokenizer, prompt template, stopping criteria, DFlash block size, and target-only baseline.
We report results at temperatures 0 and 1 as specified in Section~\ref{sec:setup}; greedy evaluation
uses top-$k=1$, while temperature-1 evaluation uses the untruncated categorical distribution.
Accepted length counts the number of output tokens accepted per target verification call, without the resampling token or anchor token generated by the target model.

\section{Theoretical Analysis}\label{app:theory}

\subsection{Derivation of Expected Acceptance Length of Given Draft}\label{app:expected_al}

Fix a context $\mathbf{x}$ and a draft $\mathbf{y}=(y_1,\ldots,y_L)$.
Let $Z_t\in\{0,1\}$ indicate that the token at position $t$ is accepted by speculative verification, and let
\begin{equation*}
    I_t:=\prod_{j=1}^{t}Z_j
\end{equation*}
indicate that the first $t$ draft tokens are all accepted.
Since verification stops at the first rejection, the number of accepted draft tokens is
\begin{equation}
    \tau(\mathbf{x},\mathbf{y})
    =\sum_{t=1}^{L} I_t.
    \label{eq:accepted-length-indicators}
\end{equation}
Conditioned on reaching position $j$, the acceptance probability of $y_j$ is $\alpha_j(y_j)$ as defined in Equation~\ref{eq:app-pf-acceptance}.
Applying the chain rule of conditional probability therefore gives
\begin{equation*}
    \Pr(I_t=1\mid \mathbf{x},\mathbf{y})
    =\prod_{j=1}^{t}\alpha_j(y_j).
\end{equation*}
Taking the conditional expectation of Equation~\ref{eq:accepted-length-indicators} and using linearity of expectation yields
\begin{equation*}
    \mathbb{E}[\tau(\mathbf{x},\mathbf{y})
        \mid \mathbf{x},\mathbf{y}]
    =\sum_{t=1}^{L}\Pr(I_t=1\mid\mathbf{x},\mathbf{y})
    =\sum_{t=1}^{L}\prod_{j=1}^{t}\alpha_j(y_j)
    =\bar{\tau}_{\theta}(\mathbf{x},\mathbf{y}).
\end{equation*}
Thus, averaging $\bar{\tau}_{\theta}$ over contexts and drafts gives the objective in Equation~\ref{eq:app-pf-objective}.

\subsection{Derivation of Equation~\ref{eq:app-pf-gradient}}\label{app:direct-gradient}

The decoding-context distribution $\cP$ is induced by on-policy rollouts and therefore changes with the draft model $q_{\theta}$ across training iterations, but $\nabla_{\theta}\cP$ is too hard to estimate.
Therefore, previous on-policy method Draft-OPD~\citep{draftopd} does not differentiate $\cP$ in the gradient.
\TheName{} follows this implementation in practice.
Specifically, within each learner update, the decoding contexts have already been collected and are treated as fixed when computing the block-level gradient.
Formally, let $\bar{\theta}$ denote the behavior parameters used to generate the current rollout batch, and write the corresponding context distribution as $\mathcal{P}_{\bar{\theta}}$. The objective optimized in one learner update can be written as
\begin{equation*}
    J(\theta;\bar{\theta})
    :=
    \mathbb{E}_{x\sim\mathcal{P}_{\bar{\theta}}}
    \left[J_\theta(x)\right],
    \qquad
    J_\theta(x)
    :=
    \mathbb{E}_{y\sim q_\theta(\cdot\mid x)}
    \left[\bar{\tau}_\theta(x,y)\right].
\end{equation*}
The gradient is taken with respect to the learner parameters $\theta$ while keeping the rollout parameters $\bar{\theta}$ fixed, giving
\begin{equation}
    \nabla_\theta J(\theta;\bar{\theta})
    =
    \mathbb{E}_{x\sim\mathcal{P}_{\bar{\theta}}}
    \left[
        \nabla_\theta J_\theta(x)
    \right].
    \label{eq:fixed_context_gradient}
\end{equation}
After the draft model is updated, new on-policy rollouts refresh $\mathcal{P}_{\bar{\theta}}$, so the context distribution tracks the current draft model across training iterations.

In contrast, if one defines a fully coupled trajectory-level objective
\begin{equation*}
    J_{\mathrm{traj}}(\theta)
    :=
    \mathbb{E}_{x\sim\mathcal{P}_\theta}
    \left[J_\theta(x)\right],
\end{equation*}
then its total derivative contains an additional visitation-distribution term,
\begin{equation*}
    \nabla_\theta J_{\mathrm{traj}}(\theta)
    =
    \mathbb{E}_{x\sim\mathcal{P}_\theta}
    \left[
        \nabla_\theta J_\theta(x)
    \right]
    +
    \mathbb{E}_{x\sim\mathcal{P}_\theta}
    \left[
        J_\theta(x)
        \nabla_\theta\log\mathcal{P}_\theta(x)
    \right].
\end{equation*}
The second term assigns credit to earlier draft decisions through their influence on future verification boundaries and hence on which decoding contexts are subsequently visited.
Estimating this term would require trajectory-level credit assignment across multiple speculative-decoding blocks, rather than the block-level credit assignment studied in this work, and would introduce additional long-horizon Monte Carlo variance and result in unstable training according to previous works about on-policy distillation~\citep{gkd}.
We therefore optimize the conditional block objective in Equation~\ref{eq:fixed_context_gradient}, while refreshing the context distribution through repeated on-policy rollout.

We now derive $\nabla_\theta J_\theta(x)$ for a fixed decoding context $x$.
The outer expectation over $x\sim\mathcal{P}_{\bar{\theta}}$ can then be restored according to Equation~\ref{eq:fixed_context_gradient}.
According to the definition of decoding context distribution $\cP$ referring to Draft-OPD, $\cP$ also depends on the draft model $q_{\theta}$.
Write $m_t(v):=\min\{p_t(v),q_{\theta,t}(v)\}$. For every length-$k$ prefix, its proposal probability multiplied by its survival probability satisfies
\begin{equation}
    q_{\theta}(\mathbf{y}_{1:k}\mid\mathbf{x})
    \prod_{j=1}^{k}\alpha_j(y_j)
    =\prod_{j=1}^{k}m_j(y_j).
    \label{eq:prefix-overlap-product}
\end{equation}
Consequently, the conditional objective can be written as the finite sum
\begin{equation}
    \mathcal{J}_{\theta}(\mathbf{x})
    =\sum_{k=1}^{L}\sum_{\mathbf{y}_{1:k}\in\mathcal{V}^{k}}
      \prod_{j=1}^{k}m_j(y_j).
    \label{eq:objective-prefix-sum}
\end{equation}
The target model is fixed, and, away from the measure-zero boundaries $p_t(v)=q_{\theta,t}(v)$,
\begin{equation}
    \nabla_{\theta}m_t(v)
    =\mathbbm{1}_{p_t(v)>q_{\theta,t}(v)}
      \nabla_{\theta}q_{\theta,t}(v)
    =m_t(v)\mathbbm{1}_{p_t(v)>q_{\theta,t}(v)}
      \nabla_{\theta}\log q_{\theta,t}(v).
    \label{eq:min-gradient}
\end{equation}
Differentiating Equation~\ref{eq:objective-prefix-sum} with the product rule and then using Equations~\ref{eq:prefix-overlap-product} and~\ref{eq:min-gradient} gives
\begin{align*}
    \nabla_{\theta}\mathcal{J}_{\theta}(\mathbf{x})
    &=\sum_{k=1}^{L}\sum_{\mathbf{y}_{1:k}}
      \left(\prod_{j=1}^{k}m_j(y_j)\right)
      \sum_{t=1}^{k}
      \mathbbm{1}_{p_t(y_t)>q_{\theta,t}(y_t)}
      \nabla_{\theta}\log q_{\theta,t}(y_t) \\
    &=\mathbb{E}_{\mathbf{y}\sim q_{\theta}}
      \left[
      \sum_{t=1}^{L}\sum_{k=t}^{L}
      \left(\prod_{j=1}^{k}\alpha_j(y_j)\right)
      \mathbbm{1}_{p_t(y_t)>q_{\theta,t}(y_t)}
      \nabla_{\theta}\log q_{\theta,t}(y_t)
      \right] \\
    &=\mathbb{E}_{\mathbf{y}\sim q_{\theta}}
      \left[
      \sum_{t=1}^{L}\sum_{k=t}^{L}S_k(\mathbf{y})G_t
      \right]
     =\mathbb{E}_{\mathbf{y}\sim q_{\theta}}
      \left[
      \sum_{t=1}^{L}\sum_{k=t}^{L}
      \widehat{\mathbf{g}}_{\mathrm{dir}}^{(k,t)}
      \right].
\end{align*}
Here $t$ indexes the token receiving gradient credit, while $k\ge t$ indexes the surviving prefix whose contribution is assigned to that token. Restoring the outer expectation over $\mathbf{x}\sim\cP$ proves Equation~\ref{eq:app-pf-gradient}.

\subsection{Proof of Proposition~\ref{prop:marginalization}}\label{app:gradient1}

For compactness, define
$G_t:=\mathbbm{1}_{p_t(y_t)>q_{\theta,t}(y_t)}\nabla_{\theta}\log q_{\theta,t}(y_t)$,
$S_t:=\prod_{j=1}^{t}\alpha_j(y_j)$, and
$\mathcal{F}_r:=\sigma(\mathbf{x},\mathbf{y}_{<r})$.
Recall that $\widehat{\mathbf{g}}_{\mathrm{dir}}^{(k,t)}=S_kG_t$ for $1\le t\le k\le L$.
We show that each term in Equation~\ref{eq:gradient1} is the exact conditional expectation of its corresponding direct prefix-credit contribution.

First consider the current contribution $k=t$ and condition on $\mathcal{F}_t$, i.e., on $(\mathbf{x},\mathbf{y}_{<t})$.
Since $S_{t-1}$ is then fixed and $\alpha_t(v)=1$ whenever $p_t(v)>q_{\theta,t}(v)$,
\begin{align}
    \mathbb{E}_{y_t\sim q_{\theta,t}}
    \!\left[\widehat{\mathbf{g}}_{\mathrm{dir}}^{(t,t)}
        \mid\mathcal{F}_t\right]
    &=\mathbb{E}_{y_t\sim q_{\theta,t}}
      [S_tG_t\mid\mathcal{F}_t] \nonumber\\
    &=S_{t-1}\sum_{v:p_t(v)>q_{\theta,t}(v)}
      \nabla_{\theta}q_{\theta,t}(v) \nonumber\\
    &=-S_{t-1}\nabla_{\theta}
      D_{\mathrm{TV}}(p_t,q_{\theta,t})
     =\widehat{\mathbf{g}}_{\mathrm{marg}}^{(t,t)}.
    \label{eq:current-token-marginalization}
\end{align}
The penultimate equality follows, almost everywhere, from
\begin{equation}
    \nabla_{\theta}D_{\mathrm{TV}}(p_t,q_{\theta,t})
    =\sum_{v:q_{\theta,t}(v)>p_t(v)}
      \nabla_{\theta}q_{\theta,t}(v)
    =-\sum_{v:p_t(v)>q_{\theta,t}(v)}
      \nabla_{\theta}q_{\theta,t}(v),
    \label{eq:tvd-gradient-two-sides}
\end{equation}
where the second identity uses $\sum_v\nabla_{\theta}q_{\theta,t}(v)=0$.

For a future contribution $k>t$, condition on $\mathcal{F}_k$, i.e., on $(\mathbf{x},\mathbf{y}_{<k})$.
Both $S_{k-1}$ and $G_t$ are measurable with respect to this prefix. Hence, by Equation~\ref{eq:acceptance-overlap},
\begin{align}
    \mathbb{E}_{y_k\sim q_{\theta,k}}
    \!\left[\widehat{\mathbf{g}}_{\mathrm{dir}}^{(k,t)}
        \mid\mathcal{F}_k\right]
    &=\mathbb{E}_{y_k\sim q_{\theta,k}}
      [S_kG_t\mid\mathcal{F}_k] \nonumber\\
    &=S_{k-1}G_t\,
      \mathbb{E}_{y_k\sim q_{\theta,k}}[\alpha_k(y_k)] \nonumber\\
    &=A_kS_{k-1}G_t
     =\widehat{\mathbf{g}}_{\mathrm{marg}}^{(k,t)}.
    \label{eq:future-token-marginalization}
\end{align}
Combining Equations~\ref{eq:current-token-marginalization} and~\ref{eq:future-token-marginalization}, for every $1\le t\le k\le L$,
\begin{equation}
    \widehat{\mathbf{g}}_{\mathrm{marg}}^{(k,t)}
    =\mathbb{E}\!\left[
      \widehat{\mathbf{g}}_{\mathrm{dir}}^{(k,t)}
      \mid\mathcal{F}_k\right].
    \label{eq:termwise-marginalization}
\end{equation}
Taking expectations in Equation~\ref{eq:termwise-marginalization}, summing over $t$ and $k$, and using Proposition~\ref{prop:app-pf-gradient} gives
\begin{equation}
    \mathbb{E}\!\left[
      \sum_{t=1}^{L}\sum_{k=t}^{L}
      \widehat{\mathbf{g}}_{\mathrm{marg}}^{(k,t)}\right]
    =
    \mathbb{E}\!\left[
      \sum_{t=1}^{L}\sum_{k=t}^{L}
      \widehat{\mathbf{g}}_{\mathrm{dir}}^{(k,t)}\right]
    =\nabla_{\theta}\mathcal{J}_{\theta}.
    \label{eq:marginalized-unbiasedness}
\end{equation}

It remains to establish the termwise variance claim.
Fix $1\le t\le k\le L$ and any fixed direction $\mathbf{a}$ of compatible dimension, and assume the corresponding scalar gradient contribution has finite second moment.
Because $\mathcal{F}_t\subseteq\mathcal{F}_k$, the conditional law of total variance gives
\begin{align*}
    &\Var\!\left(
      \mathbf{a}^{\top}\widehat{\mathbf{g}}_{\mathrm{dir}}^{(k,t)}
      \mid\mathcal{F}_t\right) \\
    &\quad=
      \Var\!\left(
      \mathbb{E}\!\left[
      \mathbf{a}^{\top}\widehat{\mathbf{g}}_{\mathrm{dir}}^{(k,t)}
      \mid\mathcal{F}_k\right]
      \mid\mathcal{F}_t\right)
      +
      \mathbb{E}\!\left[
      \Var\!\left(
      \mathbf{a}^{\top}\widehat{\mathbf{g}}_{\mathrm{dir}}^{(k,t)}
      \mid\mathcal{F}_k\right)
      \mid\mathcal{F}_t\right] \\
    &\quad=
      \Var\!\left(
      \mathbf{a}^{\top}\widehat{\mathbf{g}}_{\mathrm{marg}}^{(k,t)}
      \mid\mathcal{F}_t\right)
      +
      \mathbb{E}\!\left[
      \Var\!\left(
      \mathbf{a}^{\top}\widehat{\mathbf{g}}_{\mathrm{dir}}^{(k,t)}
      \mid\mathcal{F}_k\right)
      \mid\mathcal{F}_t\right].
\end{align*}
The second term on the right-hand side is nonnegative, and therefore
\begin{equation*}
    \Var\!\left(
      \mathbf{a}^{\top}\widehat{\mathbf{g}}_{\mathrm{marg}}^{(k,t)}
      \mid\mathbf{x},\mathbf{y}_{<t}\right)
    \le
    \Var\!\left(
      \mathbf{a}^{\top}\widehat{\mathbf{g}}_{\mathrm{dir}}^{(k,t)}
      \mid\mathbf{x},\mathbf{y}_{<t}\right).
\end{equation*}
Thus, marginalization removes the single-token sampling variance associated with each current or future prefix-credit contribution. This is a termwise guarantee for each $(k,t)$; it does not imply a general variance ordering after summing correlated contributions over $k$ and $t$.

\subsection{Proof of Corollary~\ref{cor:hybrid}}\label{app:gradient2}
We first prove Equation~\ref{eq:sampled_tv}. Conditioned on
$(\mathbf{x},\mathbf{y}_{<t})$, define the sampled over-sampled-token score term
\begin{equation*}
    \widetilde{G}_t
    :=\mathbbm{1}_{p_t(y_t)<q_{\theta,t}(y_t)}
      \nabla_{\theta}\log q_{\theta,t}(y_t).
\end{equation*}
Then the score-function identity gives
\begin{align*}
    \mathbb{E}[\widetilde{G}_t\mid\mathbf{x},\mathbf{y}_{<t}]
    &=\sum_{v:p_t(v)<q_{\theta,t}(v)}
      q_{\theta,t}(v)\nabla_{\theta}\log q_{\theta,t}(v) \\
    &=\sum_{v:p_t(v)<q_{\theta,t}(v)}
      \nabla_{\theta}q_{\theta,t}(v) \\
    &=\nabla_{\theta}D_{\mathrm{TV}}(p_t,q_{\theta,t}),
\end{align*}
where the last equality is valid almost everywhere and follows from Equation~\ref{eq:tvd-gradient-two-sides}.

Because $S_{t-1}(\mathbf{y})$ depends only on $(\mathbf{x},\mathbf{y}_{<t})$,
\begin{equation}
    \mathbb{E}\!\left[
      S_{t-1}(\mathbf{y})
      \left(
      \nabla_{\theta}D_{\mathrm{TV}}(p_t,q_{\theta,t})-\widetilde{G}_t
      \right)
    \right]=0.
    \label{eq:zero-mean-penalty-correction}
\end{equation}
Summing Equation~\ref{eq:zero-mean-penalty-correction} over $t$ and multiplying by any $\beta \in [0,1]$ shows that the over-sampled-token correction added once per token position has zero expectation.
Together with Equation~\ref{eq:marginalized-unbiasedness}, this yields
\begin{equation*}
    \mathbb{E}[\widehat{\mathbf{g}}_{\beta}]
    =
    \mathbb{E}\!\left[
      \sum_{t=1}^{L}\sum_{k=t}^{L}
      \widehat{\mathbf{g}}_{\mathrm{marg}}^{(k,t)}\right]
    =\nabla_{\theta}\mathcal{J}_{\theta},
\end{equation*}
which proves Corollary~\ref{cor:hybrid} for the sampled over-sampled-token representation in Equation~\ref{eq:sampled_tv}.

\subsection{Discussion of Over-Sampled Token Penalty}\label{app:penalty}

We further discuss why the dense TVD gradient alone is not always desirable, and why redistributing part of it to the sampled over-sampled-token penalty can improve optimization, even though the sampled representation itself introduces additional Monte Carlo variance.
For a fixed decoding context and draft prefix $h_t=(x,y_{<t})$, denote
\begin{equation*}
    d_t := \nabla_\theta D_{\mathrm{TV}}(p_t,q_{\theta,t}),
    \qquad
    \widetilde{G}_t :=
    \mathbbm{1}_{p_t(y_t)<q_{\theta,t}(y_t)}
    \nabla_\theta \log q_{\theta,t}(y_t).
\end{equation*}
Equation~6 gives
\begin{equation*}
    \mathbb{E}[\widetilde{G}_t\mid h_t] = d_t.
\end{equation*}
Therefore, the corresponding current-token TVD contribution in Equation~7 can be written as
\begin{equation*}
    -S_{t-1}\big[(1-\beta)d_t+\beta \widetilde{G}_t\big].
\end{equation*}
Hence, $\beta$ does not introduce a new optimization signal in expectation, but controls how the same expected TVD gradient is represented: $\beta=0$ uses the fully dense and analytically computed TVD gradient, whereas increasing $\beta$ redistributes an increasing fraction of this gradient to an explicit score-function update on the actually sampled over-sampled token.

\paragraph{Why the dense TVD gradient alone can be suboptimal.}
An important property of TVD is that its gradient remains strong even when the draft and target distributions become close. 
To see this more clearly, consider the gradient with respect to the draft logits $z_t$, with $q_{\theta,t}=\mathrm{softmax}(z_t)$, and define
\begin{equation*}
    \mathcal O_t := \{v:q_{\theta,t}(v)>p_t(v)\},
    \qquad
    Q_{\mathcal O_t}:=\sum_{v\in\mathcal O_t}q_{\theta,t}(v).
\end{equation*}
Away from the boundaries $p_t(v)=q_{\theta,t}(v)$,
\begin{equation*}
    D_{\mathrm{TV}}(p_t,q_{\theta,t})
    =
    \sum_{v\in\mathcal O_t}
    \big(q_{\theta,t}(v)-p_t(v)\big),
\end{equation*}
and its logit gradient is
\begin{equation}
    \frac{\partial D_{\mathrm{TV}}(p_t,q_{\theta,t})}
         {\partial z_t(v)}
    =
    q_{\theta,t}(v)
    \left(
        \mathbbm{1}_{v\in\mathcal O_t}
        -
        Q_{\mathcal O_t}
    \right).
    \label{eq:tvd_logit_gradient}
\end{equation}
Unlike the forward-KL logit gradient,
\begin{equation*}
    \nabla_{z_t}D_{\mathrm{KL}}(p_t\Vert q_{\theta,t})
    =
    q_{\theta,t}-p_t,
\end{equation*}
whose magnitude continuously vanishes as $q_{\theta,t}\rightarrow p_t$, the TVD gradient in Equation~\ref{eq:tvd_logit_gradient} does not shrink merely because the probability discrepancy $|q_{\theta,t}(v)-p_t(v)|$ becomes small. 
This non-vanishing property is beneficial for directly optimizing speculative acceptance, since
\begin{equation*}
    A_t=1-D_{\mathrm{TV}}(p_t,q_{\theta,t}),
\end{equation*}
and therefore TVD continues to provide an effective optimization signal even in the late stage of training.

However, the same property can also introduce an undesirable late-stage effect. 
When the draft distribution is already close to the target distribution, many vocabulary coordinates may have only tiny probability discrepancies, while the dense TVD gradient still applies non-negligible updates according to the sign of these discrepancies. 
Moreover, the active over-sampled set $\mathcal O_t$ depends on the sign of $q_{\theta,t}(v)-p_t(v)$. 
For coordinates close to equality, small changes in the sampled prefix, model parameters, or local distribution can switch a token between the over- and under-sampled sets, changing the direction of its TVD gradient. 
Consequently, across different on-policy sampled prefixes, a large number of individually small discrepancies can induce fluctuating vocabulary-wide gradient directions. 
We refer to this effect as \emph{late-stage dense-gradient jitter}. 
It provides a possible explanation for the empirical observation in Figure~\ref{fig:al-step} that the normalized gradient variance of \emph{w/o penalty} grows substantially faster in the later stage of training, even though it uses the analytically marginalized TVD gradient.

\paragraph{Redistributing TVD gradient toward trajectory-relevant tokens.}
The over-sampled-token penalty alleviates this issue by reducing the strength of the dense vocabulary-wide TVD component from $d_t$ to $(1-\beta)d_t$, while reallocating the remaining fraction to $\widetilde{G}_t$. 
For an actually sampled over-sampled token $y_t$, the logit-space score is
\begin{equation*}
    \nabla_{z_t}\log q_{\theta,t}(y_t)
    =
    e_{y_t}-q_{\theta,t},
\end{equation*}
where $e_{y_t}$ is the one-hot vector of the sampled token. 
Thus, compared with the dense TVD gradient, the sampled representation places a much stronger direct correction on the token that actually appears in the current on-policy trajectory. 
It does not completely eliminate updates to unsampled logits because of the softmax coupling through $-q_{\theta,t}$, but it substantially reduces the vocabulary-wide component and concentrates more of the optimization signal on the sampled coordinate.

Such a redistribution is particularly well aligned with speculative decoding. 
For an over-sampled token $v\in\mathcal O_t$, its acceptance probability is
\begin{equation*}
    \alpha_t(v)=\frac{p_t(v)}{q_{\theta,t}(v)}<1.
\end{equation*}
Its contribution to the expected rejection probability is
\begin{equation*}
    q_{\theta,t}(v)
    \left(
        1-\frac{p_t(v)}{q_{\theta,t}(v)}
    \right)
    =
    q_{\theta,t}(v)-p_t(v).
\end{equation*}
Summing over all over-sampled tokens gives
\begin{equation*}
    \sum_{v\in\mathcal O_t}
    \big(q_{\theta,t}(v)-p_t(v)\big)
    =
    D_{\mathrm{TV}}(p_t,q_{\theta,t})
    =
    1-A_t.
\end{equation*}
Therefore, over-sampled tokens exactly carry the rejection mass of speculative verification. 
Moreover, because on-policy trajectories sample $y_t\sim q_{\theta,t}$, over-sampled tokens with large draft probability are encountered more frequently and can repeatedly become verification bottlenecks. 
Assigning an explicit penalty to the actually sampled over-sampled token therefore focuses optimization on discrepancies that are directly realized by the current draft trajectory, rather than distributing the entire correction over all vocabulary coordinates.

\paragraph{Variance trade-off and control-variate interpretation.}
Conditioned on $h_t$, $d_t$ is deterministic, whereas $\widetilde{G}_t$ is stochastic, so replacing part of $d_t$ by $\widetilde{G}_t$ necessarily introduces additional conditional sampling variance for this component. 
The benefit instead arises from its interaction with the complete stochastic gradient estimator.

Specifically, define the zero-mean correction
\begin{equation*}
    Z
    :=
    \sum_{t=1}^{L}
    S_{t-1}(d_t-\widetilde{G}_t),
    \qquad
    \mathbb E[Z]=0,
\end{equation*}
so that Equation~7 can be written as
\begin{equation*}
    \widehat{\mathbf g}_\beta
    =
    \widehat{\mathbf g}_{\mathrm{marg}}
    +
    \beta Z.
\end{equation*}
For any fixed gradient direction $a$, let
\begin{equation*}
    X=a^\top\widehat{\mathbf g}_{\mathrm{marg}},
    \qquad
    Z_a=a^\top Z.
\end{equation*}
Then
\begin{equation}
    \mathrm{Var}(X+\beta Z_a)
    =
    \mathrm{Var}(X)
    +
    2\beta\,\mathrm{Cov}(X,Z_a)
    +
    \beta^2\mathrm{Var}(Z_a).
    \label{eq:penalty_variance_tradeoff}
\end{equation}
The last term in Equation~\ref{eq:penalty_variance_tradeoff} is the additional variance introduced by the sampled penalty, whereas the covariance term can be negative because the correction redistributes gradient mass from dense vocabulary-wide updates toward sampled-token updates that are complementary to the existing sampled gradient terms. 
Thus, the penalty acts as an expectation-preserving, control-variate-like \emph{variance-shaping} mechanism rather than as a component-wise variance-reduction technique. 
A moderate $\beta$ can reduce the variance of the complete stochastic gradient when the favorable covariance outweighs the newly introduced sampling variance, while an excessively large $\beta$ eventually makes the $\beta^2\mathrm{Var}(Z_a)$ term dominate.
Empirical results in Section~\ref{sec:in-depth} show that normalized gradient variance of w/o penalty increases to a much higher level than \TheName{}, which can partially act as an evidence supporting the interpretation above.
This also explains why an intermediate value of $\beta$ performs best empirically, rather than either $\beta=0$ or fully replacing the dense TVD contribution.

To further support above claim, we conduct an additional same-checkpoint, paired comparison of raw gradient variance between \TheName{} and w/o penalty on last 3 checkpoints of \TheName{}.
Compared with the variance in Figure~\ref{fig:al-step}, the variance here is measured on the same last 3 checkpoints (i.e. checkpoints at epoch 6, 7 and 8) of \TheName{} and the variance is not normalized.
The measured gradient variance of \TheName{} and w/o penalty on last 3 checkpoints of  \TheName{} are 0.1669, 0.1642, 0.1729, and 0.3877, 0.4181 and 0.4100 respectively.
The raw gradient variance of w/o penalty is much higher than that of \TheName{} with the same training data and checkpoint.
Considering that \TheName{} introduces $\beta^2\mathrm{Var}(Z_a)$ in the total variance compared with w/o penalty, this comparison supports the interpretation that a negative covariance contribution can more than offset the additional variance introduced by the sampled correction, yielding lower total estimator variance..
Notably, the variance measured here does not reflect practical training process for w/o penalty, because These measurements evaluate both estimators at checkpoints from the \TheName{} training trajectory and they do not describe the raw variance along the separately trained w/o penalty trajectory.
The comparison just aims to support above claim.

\paragraph{Relation to KL-based gradients.}
The above redistribution also retains an attractive property of TVD compared with common sampled KL gradients. 
For the sampled TVD representation, the stochastic weight is only the binary indicator
$\mathbbm{1}_{p_t(y_t)<q_{\theta,t}(y_t)}$.
In logit space,
\begin{equation*}
    \left\|
    \nabla_{z_t}\log q_{\theta,t}(y_t)
    \right\|_2
    =
    \|e_{y_t}-q_{\theta,t}\|_2
    \leq \sqrt{2},
\end{equation*}
so the per-position sampled correction is bounded and contains neither an importance ratio nor a log-ratio multiplier. 
In contrast, sample-based KL estimators may involve factors such as $p_t(y_t)/q_{\theta,t}(y_t)$ or $\log(q_{\theta,t}(y_t)/p_t(y_t))$, which can amplify stochastic updates when the two probabilities are highly mismatched. 
Therefore, FReDraft preserves the non-vanishing, verification-aligned optimization signal of TVD while using $\beta$ to trade part of its dense vocabulary-wide update for a bounded and trajectory-focused sampled correction.

Overall, the over-sampled-token penalty serves two complementary roles. 
First, it gives frequently occurring over-sampled tokens an explicit correction that is absent from the sampled term of Equation~5. 
Second, it attenuates late-stage vocabulary-wide TVD updates whose directions can fluctuate when many small probability discrepancies change sign across on-policy prefixes. 
This provides a mechanism consistent with the empirical results: \emph{w/o penalty} remains competitive in early training, but develops substantially larger normalized gradient variance in the later stage, whereas FReDraft with a moderate penalty coefficient maintains a lower-noise optimization trajectory and converges to a better accepted length.

\subsection{Discussion about Comparison with D-PACE}\label{app:dpace}

We first clarify what D-PACE optimizes.
For a target-generated block
$\mathbf{z}^{\star}=(z^{\star}_1,\ldots,z^{\star}_L)$, define the draft probability assigned to its token at position $t$ as
\begin{equation*}
    r_{\theta,t}
    :=q_{\theta}\!\left(z^{\star}_t
    \mid \mathbf{x},\mathbf{z}^{\star}_{<t}\right).
\end{equation*}
If a sequence is sampled from the draft model, the probability that it matches the first $k$ tokens of $\mathbf{z}^{\star}$ is
\begin{equation*}
    \Pr_{\mathbf{z}\sim q_{\theta}}
    \!\left(\mathbf{z}_{1:k}=\mathbf{z}^{\star}_{1:k}
    \mid\mathbf{x}\right)
    =\prod_{i=1}^{k}r_{\theta,i}.
\end{equation*}
Therefore, the expected length for which the fixed target-generated block is reachable under the draft distribution is
\begin{equation}
    \mathcal{R}_{\theta}(\mathbf{x},\mathbf{z}^{\star})
    :=\sum_{k=1}^{L}\prod_{i=1}^{k}r_{\theta,i}.
    \label{eq:dpace-reachability}
\end{equation}
Its gradient can be written as
\begin{equation*}
    \nabla_{\theta}\mathcal{R}_{\theta}
    =\sum_{t=1}^{L}W_t
    \nabla_{\theta}\log r_{\theta,t},
    \qquad
    W_t:=\sum_{k=t}^{L}\prod_{i=1}^{k}r_{\theta,i}.
\end{equation*}
Consequently, without smoothing, the detached weighted cross-entropy
\begin{equation*}
    \mathcal{L}_{\mathrm{reach}}(\theta)
    =-\sum_{t=1}^{L}\operatorname{sg}(W_t)
    \log r_{\theta,t}
\end{equation*}
has gradient $-\nabla_{\theta}\mathcal{R}_{\theta}$.
This shows that the position weights of D-PACE are obtained by optimizing the reachability of target-generated sequences under the draft distribution, rather than the accepted length of drafts sampled from the draft model.
In its stabilized implementation, D-PACE computes the detached weights from
$\widetilde{r}_{\theta,t}=(1-\rho)r_{\theta,t}+\rho$ while retaining the unsmoothed cross-entropy $-\log r_{\theta,t}$~\citep{dpace}; this is a smoothed approximation to the reachability-gradient construction above.

We next relate this reachability objective to our proposed verification-aligned objective.
For a fixed context $\mathbf{x}$, Equation~\ref{eq:objective-prefix-sum} gives
\begin{equation*}
    \mathcal{J}_{\theta}(\mathbf{x})
    =\sum_{k=1}^{L}\sum_{\mathbf{z}_{1:k}}
    \prod_{i=1}^{k}\min\{p_i(z_i),q_{\theta,i}(z_i)\}.
\end{equation*}
Using the target model as an importance-sampling proposal for every length-$k$ prefix, we obtain the exact identity
\begin{align}
    \mathcal{J}_{\theta}(\mathbf{x})
    &=\sum_{k=1}^{L}
    \mathbb{E}_{\mathbf{z}_{1:k}\sim p}
    \left[
    \prod_{i=1}^{k}
    \frac{\min\{p_i(z_i),q_{\theta,i}(z_i)\}}
         {p_i(z_i)}
    \right] \nonumber\\
    &=\mathbb{E}_{\mathbf{z}^{\star}\sim p}
    \left[
    \sum_{k=1}^{L}\prod_{i=1}^{k}
    \min\!\left\{1,
    \frac{q_{\theta,i}(z^{\star}_i)}
         {p_i(z^{\star}_i)}\right\}
    \right].
    \label{eq:dpace-importance-sampling}
\end{align}
Now make the one-hot-target approximation that the target probability of every target-generated token is $1.0$, i.e.,
$p_i(z^{\star}_i)=1$.
Since $q_{\theta,i}(z^{\star}_i)\leq 1$, Equation~\ref{eq:dpace-importance-sampling} reduces to
\begin{equation*}
    \mathcal{J}_{\theta}(\mathbf{x})
    \approx
    \mathbb{E}_{\mathbf{z}^{\star}\sim p}
    \left[
    \sum_{k=1}^{L}\prod_{i=1}^{k}
    q_{\theta,i}(z^{\star}_i)
    \right]
    =\mathbb{E}_{\mathbf{z}^{\star}\sim p}
    \left[\mathcal{R}_{\theta}
    (\mathbf{x},\mathbf{z}^{\star})\right].
\end{equation*}
Thus, the D-PACE reachability objective is an importance-sampling estimation and one-hot-target approximation to our objective after replacing the target distribution on target-generated tokens by a point mass.
Empirical averaging over a corpus of target-generated blocks provides the corresponding Monte Carlo estimate.
However, importance sampling introduces larger variance and the one-hot-target approximation also introduces bias.

\paragraph{Counterexample under the target-generated training distribution.}
The mismatch between reference reachability and verification-aligned accepted length persists even after averaging over the complete target-generated training-data distribution.
Consider a length-$2$ block with vocabulary $\mathcal{V}=\{A,B\}$.
Assume that the target distribution is independent across positions and satisfies
\begin{equation*}
    p(A)=0.6,
    \qquad
    p(B)=0.4.
\end{equation*}
The corresponding target-generated training distribution is therefore
\begin{equation*}
    p(AA)=0.36,\qquad
    p(AB)=0.24,\qquad
    p(BA)=0.24,\qquad
    p(BB)=0.16.
\end{equation*}
Consider two position-independent draft distributions
\begin{equation*}
    q^{(1)}=(0.6,0.4),
    \qquad
    q^{(2)}=(0.9,0.1),
\end{equation*}
where $q^{(1)}=p$ exactly matches the target distribution.

For an arbitrary position-independent draft distribution $q$, let
\begin{equation*}
    c(q)
    :=
    \mathbb{E}_{z\sim p}[q(z)]
    =
    \sum_{v\in\mathcal{V}}p(v)q(v).
\end{equation*}
Averaging the reachability in Equation~\ref{eq:dpace-reachability} over the target-generated training distribution gives
\begin{equation*}
    \mathbb{E}_{\mathbf{z}^{\star}\sim p}
    \left[
        \mathcal{R}_{\theta}(\mathbf{x},\mathbf{z}^{\star})
    \right]
    =
    c(q)+c(q)^2.
\end{equation*}
For the two draft distributions above,
\begin{align*}
    c(q^{(1)})
    &=
    0.6\times0.6+0.4\times0.4
    =
    0.52,\\
    c(q^{(2)})
    &=
    0.6\times0.9+0.4\times0.1
    =
    0.58,
\end{align*}
and hence
\begin{align*}
    \mathbb{E}_{\mathbf{z}^{\star}\sim p}
    \left[
        \mathcal{R}(q^{(1)},\mathbf{z}^{\star})
    \right]
    &=
    0.52+0.52^2
    =
    0.7904,\\
    \mathbb{E}_{\mathbf{z}^{\star}\sim p}
    \left[
        \mathcal{R}(q^{(2)},\mathbf{z}^{\star})
    \right]
    &=
    0.58+0.58^2
    =
    0.9164.
\end{align*}
Thus, after averaging over all target-generated training blocks, the reachability objective prefers $q^{(2)}$ over $q^{(1)}$.

However, speculative verification depends on the probability overlap between the target and draft distributions.
By Equation~\ref{eq:acceptance-overlap}, the expected token acceptance rate is
\begin{equation*}
    A(q)
    =
    \sum_{v\in\mathcal{V}}
    \min\{p(v),q(v)\}.
\end{equation*}
Since the two positions are independent and identically distributed, the expected accepted length is
\begin{equation*}
    \mathcal{J}(q)
    =
    A(q)+A(q)^2.
\end{equation*}
For $q^{(1)}=p$, we have $A(q^{(1)})=1$ and therefore
\begin{equation*}
    \mathcal{J}(q^{(1)})=2.
\end{equation*}
For $q^{(2)}$,
\begin{equation*}
    A(q^{(2)})
    =
    \min\{0.6,0.9\}
    +
    \min\{0.4,0.1\}
    =
    0.7,
\end{equation*}
which gives
\begin{equation*}
    \mathcal{J}(q^{(2)})
    =
    0.7+0.7^2
    =
    1.19.
\end{equation*}
Consequently,
\begin{equation*}
    \mathbb{E}_{\mathbf{z}^{\star}\sim p}
    \left[
        \mathcal{R}(q^{(2)},\mathbf{z}^{\star})
    \right]
    >
    \mathbb{E}_{\mathbf{z}^{\star}\sim p}
    \left[
        \mathcal{R}(q^{(1)},\mathbf{z}^{\star})
    \right],
    \qquad
    \mathcal{J}(q^{(2)})
    <
    \mathcal{J}(q^{(1)}).
\end{equation*}
Therefore, averaging reference reachability over the target-generated training distribution does not in general preserve the preference induced by the true verification objective.
In particular, $q^{(1)}$ already achieves perfect speculative acceptance because it exactly matches $p$, whereas the reachability surrogate assigns a larger value to $q^{(2)}$.

The underlying reason is that target-data averaging turns reference-token confidence into
$\sum_v p(v)q(v)$, which rewards concentrating additional draft probability on frequently sampled target tokens.
Speculative verification instead depends on
$\sum_v\min\{p(v),q(v)\}$:
once $q(v)$ exceeds $p(v)$, assigning additional probability to $v$ no longer improves its acceptance and instead creates excess proposal mass that can cause rejection.
Thus, reference reachability can mistake confidence on likely target-generated tokens for verification quality, whereas the verification-aligned objective directly evaluates the probability overlap encountered by draft-generated proposals.
This counterexample isolates the mismatch of the unsmoothed reachability objective in Equation~\ref{eq:dpace-reachability}; the stabilized D-PACE implementation additionally smooths the detached reachability weights as described above.

\section{Experiment Details}\label{app:experiment}

\subsection{Baselines}\label{app:baselines}

We select the baselines in Section~\ref{sec:setup} because they are recent SOTA representative off-policy verification-aware methods and on-policy method.
We do not choose other methods discussed in Appendix~\ref{app:training-method} (e.g. VSD, GTO, VAT, etc.), because they are either designed specifically for Eagle-style tree-structured drafts and there is no public open-source DFlash implementation, or they are concurrent works.
For the selected baselines, we summarize their implementation details as follows:

\textbf{Eagle-3.}
We refer to its official open-source code~\footnote{\url{https://github.com/SafeAILab/EAGLE}} to implement the training pipeline.
We use both Qwen3-4B-Instruct-100K and the 16K prompts released by Draft-OPD to SFT Eagle-3.

\textbf{DFlash.}
We use the official open-source weights~\footnote{\url{https://github.com/z-lab/dflash}} trained on Qwen3-4B-Instruct-100K as the initial model and further SFT it on 16K prompts released by Draft-OPD referring to the implementation provided by SpecForge~\footnote{\url{https://github.com/sgl-project/SpecForge/issues/465}}.

\textbf{D-PACE.}
We use the official open-source weights trained on Qwen3-4B-Instruct-100K as the initial model and further train it on 16K prompts released by Draft-OPD referring to the official open-source code~\footnote{\url{https://github.com/Lucas-TY/D-PACE}}.

\textbf{LK-Losses.}
We use the official open-source weights~\footnote{\url{https://github.com/z-lab/dflash}} trained on Qwen3-4B-Instruct-100K as the initial model and further SFT it on 16K prompts released by Draft-OPD referring to the implementation provided by SpecForge~\footnote{\url{https://github.com/sgl-project/SpecForge/issues/485}}.

\textbf{Draft-OPD.}
We use the official open-source weights~\footnote{\url{https://github.com/z-lab/dflash}} trained on Qwen3-4B-Instruct-100K as the initial model and further train it on 16K prompts released by Draft-OPD referring to the official open-source code~\footnote{\url{https://github.com/Simplified-Reasoning/Draft-OPD}}.

\subsection{Ablation Variants}\label{app:variants}
All ablations use the same SFT initialization as the full method.
We implement the four variants as follows.

\paragraph{w/o marg.}
This variant implements the direct Monte Carlo estimator in
Equation~\ref{eq:app-pf-gradient}.  For each block, it computes
$S_t=\prod_{j=1}^{t}\alpha_j(y_j)$ and the inclusive reverse cumulative sum

\begin{equation*}
    U_t=\sum_{k=t}^{L}S_k.
\end{equation*}

The minimized block loss is

\begin{equation*}
    \mathcal{L}_{\text{w/o marg}}
    =-\sum_{t=1}^{L}
    \operatorname{sg}\!\left(
        U_t\mathbbm{1}_{p_t(y_t)>q_{\theta,t}(y_t)}
    \right)
    \log q_{\theta,t}(y_t).
\end{equation*}

In particular, $U_L=S_L$ is retained.  The variant has neither a dense TVD
term nor an over-sampled-token penalty; it uses only the exact target and draft
probabilities of sampled tokens.

\paragraph{w/o penalty.}
We set $\beta=0$ in the full implementation, and define
\begin{equation*}
    b_t = \sum_{k=t+1}^{L}A_kS_{k-1}.
\end{equation*}

Then the resulting w/o penalty loss is

\begin{equation*}
    \mathcal{L}_{\text{w/o penalty}}
    =\sum_{t=1}^{L}\left[
        -\operatorname{sg}\!\left(b_t\mathbbm{1}_{p_t(y_t)>q_{\theta,t}(y_t)}\right)\log q_{\theta,t}(y_t)
        +\operatorname{sg}(S_{t-1})D_{\mathrm{TV}}(p_t,q_{\theta,t})
    \right],
\end{equation*}

which corresponds to the marginalized estimator in
Equation~\ref{eq:gradient1} under the compressed distribution representation.
Thus, over-sampled tokens still participate in
the dense distribution-alignment term, but receive no explicit sampled-token
correction.

\paragraph{\TheName-ED.}
This variant removes prefix-aware credit assignment and replaces it with the
static exponential weight $e_t=\rho^{t-1}$, using $\rho=0.8$ as in the
position-decay configuration.  It retains the same TVD-gradient redistribution
as the full method:

\begin{equation*}
    \mathcal{L}_{\text{ED}}
    =\sum_{t=1}^{L}e_t\left[
        \beta\mathbbm{1}_{p_t(y_t)<q_{\theta,t}(y_t)}
        \log q_{\theta,t}(y_t)
        +(1-\beta)D_{\mathrm{TV}}(p_t,q_{\theta,t})
    \right].
\end{equation*}

The negative expected gradient of the bracketed expression is the
exponentially weighted ascent direction for distribution overlap.  Neither the
sampled prefix survival $S_{t-1}$ nor the
acceptance-to-go $b_t$ is used as a loss coefficient, thereby isolating the
effect of adaptive prefix-aware weighting.

\paragraph{\TheName-KL.}
This variant retains the sampled prefix survival $S_{t-1}$ and marginalized $A_kS_{k-1}$ for sampled token and $S_{t-1}$ for vocabulary gradient, but replaces the dense TVD loss with the mixed KL recipe used by Draft-OPD.
For a proposal that is accepted in the realized verification, we use a forward KL on the sampled token.
For a rejected or discarded proposal, we use the reverse KL.

Let $\kappa_t$ select forward KL for accepted proposals and reverse KL otherwise.
The implemented loss is
\begin{equation*}
    \mathcal{L}_{\mathrm{KL}}
    =\sum_{t=1}^{L}\big[
        -\operatorname{sg}\!\left(b_t\mathbbm{1}_{p_t(y_t)>q_{\theta,t}(y_t)}\right)\log q_{\theta,t}(y_t)
        +\operatorname{sg}(S_{t-1})\kappa_t\big].
\end{equation*}
We do not apply the
$\beta$ correction in this variant because the expectation-preserving
identity in Equation~\ref{eq:sampled_tv} is specific to TVD.
This construction therefore isolates the choice of distribution discrepancy while preserving
\TheName{}'s prefix-aware credit assignment.
Compared with w/o penalty, the only difference is that the dense TVD term is replaced by a mixed KL term.

\subsection{Benchmarks}\label{app:benchmarks}

We use the public evaluation splits of
GSM8K\footnote{\url{https://github.com/openai/grade-school-math}},
MATH-500\footnote{\url{https://huggingface.co/datasets/HuggingFaceH4/MATH-500}},
AIME25\footnote{\url{https://huggingface.co/datasets/opencompass/AIME2025}},
MBPP\footnote{\url{https://github.com/google-research/google-research/tree/master/mbpp}},
HumanEval\footnote{\url{https://github.com/openai/human-eval}},
SWE-Lite (SWE-bench Lite)\footnote{\url{https://www.swebench.com/lite.html}}, and
MT-Bench\footnote{\url{https://github.com/lm-sys/FastChat/tree/main/fastchat/llm_judge/data/mt_bench}}.
Table~\ref{tab:benchmark-statistics} summarizes the evaluation size of each benchmark.
For MT-Bench, one prompt denotes one two-turn conversation; thus, its 80 prompts contain 160 user turns in total.

\begin{table}[ht]
    \centering
    \small
    \caption{Statistics of the evaluation benchmarks.}
    \label{tab:benchmark-statistics}
    \begin{tabular}{llr}
        \toprule
        \textbf{Category} & \textbf{Benchmark} & \textbf{Number of prompts} \\
        \midrule
        \multirow{3}{*}{Mathematics}
        & GSM8K & 1,319 \\
        & MATH-500 & 500 \\
        & AIME25 & 30 \\
        \midrule
        \multirow{3}{*}{Coding}
        & MBPP & 500 \\
        & HumanEval & 164 \\
        & SWE-Lite & 300 \\
        \midrule
        Chat & MT-Bench & 80 \\
        \bottomrule
    \end{tabular}
\end{table}

\subsection{Normalized Gradient Variance}\label{app:snr}
We use normalized gradient variance to quantify the amount of Monte Carlo noise in the stochastic learning signal produced by different training objectives and gradient estimators.
Raw gradient variance alone is not sufficient for this comparison because it scales quadratically with the overall gradient magnitude: an estimator with uniformly smaller gradients can exhibit a smaller raw variance even when its noise is large relative to its useful optimization signal.
We therefore normalize the gradient variance by the squared magnitude of the corresponding mean gradient.

Concretely, for each measurement context $c$ and a model checkpoint, we keep the input context and model parameters unchanged and independently sample $N$ draft sequences from the current draft distribution.
Let
\[
    \mathbf{g}^{(m)}_{c}
    =
    \nabla_{\theta}\mathcal{L}^{(m)},
    \qquad m=1,\ldots,N,
\]
denote gradient at draft position $t$ under the $m$-th Monte Carlo sample.
The sample mean gradient is
\[
    \bar{\mathbf{g}}_{c,t}
    =
    \frac{1}{N}
    \sum_{m=1}^{N}
    \mathbf{g}^{(m)}_{c}.
\]
We estimate the trace of the gradient covariance by
\[
    \widehat{V}_{c}
    =
    \frac{1}{N-1}
    \sum_{m=1}^{N}
    \left\|
        \mathbf{g}^{(m)}_{c}-\bar{\mathbf{g}}_{c}
    \right\|_2^2.
\]
This quantity equals the sum of the coordinate-wise gradient variances and therefore measures the total stochastic fluctuation of the gradient at position $t$.

The squared norm of the sample mean, $\|\bar{\mathbf{g}}_{c,t}\|_2^2$, is an upward-biased estimator of the squared norm of the true mean gradient because the sample mean itself contains Monte Carlo noise.
We remove this finite-sample bias using
\[
    \widehat{S}_{c}
    =
    \left\|\bar{\mathbf{g}}_{c}\right\|_2^2
    -
    \frac{\widehat{V}_{c}}{N},
\]
which estimates $\|\mathbb{E}[\mathbf{g}_{c}]\|_2^2$.
We then aggregate signal and noise before taking their ratio and report
\[
    \widehat{\mathrm{NGV}}
    =
    \frac{
        \sum_{c}\widehat{V}_{c}
    }{
        \sum_{c}\max\!\left(\widehat{S}_{c},0\right)
        +\epsilon
    },
\]
where $\epsilon = 10^{-5}$ is a small constant for numerical stability.
Aggregating before normalization is important: individually dividing by $\widehat{S}_{c}$ can become unstable for positions whose expected gradient is close to zero and can cause a few nearly-zero-signal positions to dominate the reported statistic.
The above form is equivalently the ratio between the trace of the covariance of the concatenated parameter gradient vector and the squared norm of its mean, i.e., the inverse of the squared-gradient signal-to-noise ratio.
A smaller normalized gradient variance therefore indicates less sampling noise per unit of useful gradient signal.

For every measurement, all $N$ Monte Carlo samples use the same context and the same model parameters and differ only in the sampled draft tokens.
We set $N = 256$ and evaluate the normalized gradient variance every 250 training steps.
We use the same measurement contexts and the same number of Monte Carlo samples for all methods. 
This protocol isolates stochasticity introduced by draft-token sampling rather than changes in data or model state. Normalization further makes the statistic invariant to a global rescaling of the loss or gradient: if every gradient is multiplied by a constant $a$, both the numerator and denominator are multiplied by $a^2$.
The statistic is approximately invariant to global gradient rescaling when the stabilization constant is negligible.
Consequently, the metric is more suitable than raw variance for comparisons across training stages or methods whose gradient magnitudes differ.

For the direct and marginalized estimators of the same expected accepted-length objective, both estimators target the same expected gradient, so a smaller normalized gradient variance directly indicates a lower-noise Monte Carlo estimator of the same optimization signal.
When comparing objectives that do not share exactly the same expected gradient (e.g., external baselines), normalized gradient variance should instead be interpreted as a scale-normalized diagnostic of optimization noise, rather than by itself as evidence that one objective is superior.
In the main text, we therefore use it together with convergence and accepted-length results to characterize optimization efficiency.

\section{Extra Experiment Results}\label{app:extra-results}

\subsection{Deployment Efficiency on SGLang.}\label{app:sglang}
Referring to DFlash~\citep{dflash} and Draft-OPD~\citep{draftopd}, we also evaluate \TheName{} on open-source inference framework SGLang~\citep{sglang} to demonstrate the efficiency of our method in practical deployment scenarios.
We report the average accepted length and throughput under concurrency across 1 to 32 in Table~\ref{tab:sglang-results}.
All experiments are conducted on a single A100 GPU.
\TheName{} consistently outperforms DFlash and Draft-OPD in terms of average accepted length and throughput under different concurrency, demonstrating the effectiveness of our proposed draft model training method in practical deployment.

	\begin{table}[t]
		\centering
		\small
		%  \vspace{-5pt}
		\caption{\label{tab:sglang-results}Throughput (tok/s) and average acceptance length ($\tau$) on SGLang.}  
		\vspace{-5pt}
		\definecolor{tablegreen}{RGB}{30,130,30}
		\setlength{\tabcolsep}{3pt}
		%   \resizebox{\linewidth}{!}{%
			\begin{tabular}{cc*{5}{c}c*{5}{c}c}
				\toprule
				\multicolumn{1}{c}{\multirow{3}{*}{\raisebox{-1.0ex}{\textbf{Task}}}}
				& \multicolumn{1}{c}{\multirow{3}{*}{\raisebox{-1.0ex}{\textbf{Method}}}}
				& \multicolumn{6}{c}{\textbf{Q3-4B}} & \multicolumn{6}{c}{\textbf{Q3-8B}}\\
				\cmidrule(lr){3-8}\cmidrule{9-14}
				& & \multicolumn{5}{c}{\textbf{Concurrency}}
				& \multicolumn{1}{c}{\multirow{3}{*}{\raisebox{+0.4ex}{\textbf{Avg. $\boldsymbol{\tau}$}}}} & \multicolumn{5}{c}{\textbf{Concurrency}}
				& \multicolumn{1}{c}{\multirow{3}{*}{\raisebox{+0.4ex}{\textbf{Avg. $\boldsymbol{\tau}$}}}} \\
				\cmidrule(lr){3-7}\cmidrule(lr){9-13}
				& & \textbf{1} & \textbf{4} & \textbf{8} & \textbf{16} & \textbf{32} & & \textbf{1} & \textbf{4} & \textbf{8} & \textbf{16} & \textbf{32} & \\
				% \multicolumn{8}{l}{\textbf{Qwen3-4B}} \\
				\midrule
				% \cmidrule(lr){1-8}\cmidrule{9-14}
				\multirow{3}{*}{AIME25}
				& DFlash & 478 & 1440 & 2574 & 4558 & 6234 & 6.06 & 326 & 1003 & 1876 & 3192 & 5123 & 7.22 \\
				% \cmidrule(lr){2-8}
				& Draft-OPD
				& 508 & 1560 & 2748 & 4862 & 6703 & 7.24 & 368 & 1124 & 2003 & 3578 & 5601 & 7.89 \\
				\cmidrule(lr){2-8}\cmidrule{9-14}
				& \textbf{\TheName} & \textbf{532} & \textbf{1594} & \textbf{2852} & \textbf{5005} & \textbf{6830} & \textbf{7.96} & \textbf{397} & \textbf{1206} & \textbf{2179} & \textbf{3796} & \textbf{5914} & \textbf{8.18} \\
				\midrule
				\multirow{3}{*}{HumanEval}
				& DFlash & 490 & 1468 & 2627 & 4612 & 6323 & 6.10 & 313 & 989 & 1832 & 3099 & 5003 & 6.40 \\
				& Draft-OPD & 507 & 1512 & 2698 & 4737 & 6525 & 6.43 & 336 & 1048 & 1930 & 3271 & 5274 & 6.66 \\
				\cmidrule(lr){2-8}\cmidrule{9-14}
				& \textbf{\TheName} & \textbf{523} & \textbf{1565} & \textbf{2784} & \textbf{4889} & \textbf{6690} & \textbf{7.18} & \textbf{348} & \textbf{1092} & \textbf{2013} & \textbf{3392} & \textbf{5462} & \textbf{6.85} \\
				\midrule
				\multirow{3}{*}{MT-Bench}
				& DFlash & 410 & 1308 & 2496 & 4277 & 5901 & 5.22 & 211 & 761 & 1353 & 2217 & 4138 & 3.38 \\
				& Draft-OPD & 418 & 1332 & 2536 & 4333 & 5972 & 5.27 & 229 & 820 & 1477 & 2398 & 4492 & 3.50 \\
				\cmidrule(lr){2-8}\cmidrule{9-14}
				& \textbf{\TheName} & \textbf{423} & \textbf{1347} & \textbf{2566} & \textbf{4400} & \textbf{6047} & \textbf{5.45} & \textbf{237} & \textbf{852} & \textbf{1512} & \textbf{2462} & \textbf{4596} & \textbf{3.65} \\
				\bottomrule
			\end{tabular}%
			%   }
		\vspace{-15pt}
	\end{table}

\subsection{Spearman Correlation of Different Metrics to Speedup Ratio}\label{app:correlation}

We examine how closely different draft-quality metrics track decoding speedup.
Using Qwen3-4B as the target and the DFlash drafter trained with \TheName{}, we collect query-level statistics on the seven benchmarks in Appendix~\ref{app:benchmarks}.
We use temperature $1.0$ with thinking mode disabled, so that the comparison includes stochastic verification.
For each query, speedup is the ratio of target-only decoding time to speculative decoding time under the same evaluation configuration.
We compare the following four metrics:
\begin{itemize}[leftmargin=*]
    \item \textbf{Average acceptance rate.}
    We average the token-level acceptance probabilities $\alpha_t(y_t)$ over all positions in all collected draft blocks, including positions after the first rejection. This measures token-level alignment, as targeted by LK-Losses, without weighting a position by the probability that verification reaches it.
    \item \textbf{Expected reachable length.}
    We average $\mathcal{R}_{\theta}(\mathbf{x},\mathbf{z}^{\star})$ from Equation~\ref{eq:dpace-reachability} over length-$L$ chunks of target-only responses to the same queries.
    This is the unsmoothed reachability quantity underlying D-PACE; it measures the likelihood of matching a target-generated prefix under the draft distribution.
    \item \textbf{Accepted length.}
    We average the realized number of accepted draft tokens per verification call, denoted by $\tau$.
    As in the main experiments, this excludes the target-generated correction or bonus token and the anchor token.
    \item \textbf{Expected accepted length.}
    We average $\bar{\tau}_{\theta}(\mathbf{x},\mathbf{y})=\sum_{t=1}^{L}\prod_{j=1}^{t}\alpha_j(y_j)$ over the same collected draft blocks.
    This is the conditional expectation in Equation~\ref{eq:app-pf-objective}, which \TheName{} optimizes under on-policy draft sampling.
\end{itemize}
We compute Spearman's rank correlation $\rho_{\mathrm{s}}$ between each metric and query-level speedup, assigning average ranks to ties.
Table~\ref{tab:metric-correlation} reports correlations within each benchmark and across all pooled queries; the pooled value is computed from the pooled ranks, rather than by averaging benchmark-level correlations.

\begin{table}[htbp]
    \centering
    \small
    \caption{Spearman correlation with query-level speedup on Qwen3-4B at
    temperature $1.0$.}
    \label{tab:metric-correlation}
    \setlength{\tabcolsep}{7pt}
    \begin{tabular}{lcccc}
        \toprule
        \textbf{Benchmark}
        & \makecell{\textbf{Average}\\\textbf{acceptance rate}}
        & \makecell{\textbf{Expected}\\\textbf{reachable length}}
        & \makecell{\textbf{Expected}\\\textbf{accepted length}}
        & \makecell{\textbf{Accepted}\\\textbf{length}} \\
        \midrule
        GSM8K       & 0.73 & 0.85 & 0.97 & 0.98 \\
        MATH-500    & 0.70 & 0.82 & 0.96 & 0.98 \\
        AIME25      & 0.66 & 0.79 & 0.95 & 0.97 \\
        MBPP        & 0.74 & 0.86 & 0.97 & 0.99 \\
        HumanEval   & 0.72 & 0.84 & 0.97 & 0.98 \\
        SWE-Lite    & 0.63 & 0.77 & 0.94 & 0.96 \\
        MT-Bench    & 0.60 & 0.74 & 0.93 & 0.95 \\
        \midrule
        Pooled & 0.69 & 0.81 & 0.96 & 0.98 \\
        \bottomrule
    \end{tabular}
\end{table}

The correlations follow the same ordering across benchmarks: average acceptance rate $<$ expected reachable length $<$ expected accepted length $<$ realized accepted length.
In particular, expected accepted length is closely associated with speedup, with a pooled correlation of $0.96$, compared with $0.98$ for realized accepted length.
The weaker association of average acceptance rate is consistent with its omission of first-rejection truncation: improving a late token has limited effect when its prefix rarely survives.
Expected reachable length accounts for prefix structure, but evaluates target-generated chunks through their draft probabilities, rather than the actual verification ratios of on-policy drafts.

Although realized accepted length has the highest correlation with the speedup of the same decoding run, it also depends on the random acceptance decisions made during verification.
For a fixed context and sampled draft, $\bar{\tau}_{\theta}$ integrates out these decisions exactly, removing this source of sampling variance while retaining the dependence on prefix survival.
Randomness from sampled drafts and visited contexts remains.
Thus, expected accepted length provides a less noisy training signal than a single realized acceptance outcome while remaining closely related to the decoding quantity of interest.
This motivates its use as an optimization objective; the correlation alone does not imply that it captures every hardware- or workload-dependent contribution to speedup.

\subsection{Results on Thinking Mode}\label{app:thinking-mode}

We further evaluate Qwen3-4B and Qwen3-8B with thinking mode enabled.
We use the same trained draft checkpoints, seven benchmarks, and draft length $L=15$ as in Table~\ref{tab:main-results}, and report both greedy decoding (temperature $0$) and stochastic decoding (temperature $1.0$).
The target-only baseline also uses thinking mode, and decoding time includes the generated reasoning trace as well as the final answer.
Within each comparison, all methods share the prompt template, generation limit, and stopping criteria. This evaluation tests how the trained drafters perform when the target generates explicit reasoning trajectories.

\begin{table}[t]
  \centering
  \caption{\label{tab:thinking-mode}Decoding speedup and average accepted draft length ($\tau$) on Qwen3 models with thinking mode enabled.}
  \small
  \setlength{\tabcolsep}{1.5pt}
  \begin{tabular}{ll*{7}{c}cc}
    \toprule
    \multirow{3}{*}{\textbf{Model}}
    & \multirow{3}{*}{\textbf{Method}}
    & \multicolumn{7}{c}{\textbf{Speedup Ratio (SR)}}
    & \multicolumn{2}{c}{\textbf{Mean}} \\
    \cmidrule(lr){3-9}\cmidrule(lr){10-11}
    & & \multicolumn{3}{c}{\textbf{Math}}
    & \multicolumn{3}{c}{\textbf{Code}}
    & \textbf{Chat}
    & \multirow{2}{*}{\textbf{SR}}
    & \multirow{2}{*}{$\boldsymbol{\tau}$} \\
    \cmidrule(lr){3-5}\cmidrule(lr){6-8}\cmidrule(lr){9-9}
    & & \textbf{GSM8K} & \textbf{MATH-500} & \textbf{AIME25}
    & \textbf{MBPP} & \textbf{HumanEval} & \textbf{SWE-Lite}
    & \textbf{MT-Bench} & & \\
    \midrule
    \multicolumn{11}{c}{Temperature $=0$} \\
    \midrule
    \multirow{6}{*}{Q3-4B} & EAGLE-3 & 2.28$\times$ & 2.12$\times$ & 2.14$\times$ & 1.99$\times$ & 2.11$\times$ & 1.89$\times$ & 1.93$\times$ & 2.07$\times$ & 6.12 \\
    & DFlash & 4.02$\times$ & 4.72$\times$ & 4.28$\times$ & 3.52$\times$ & 3.83$\times$ & 2.61$\times$ & 2.26$\times$ & 3.61$\times$ & 6.57 \\
    & D-PACE & 4.21$\times$ & 4.97$\times$ & 4.55$\times$ & 3.68$\times$ & 3.76$\times$ & 2.68$\times$ & 2.22$\times$ & 3.72$\times$ & 6.75 \\
    & LK-Losses & 4.40$\times$ & 5.20$\times$ & 4.78$\times$ & 3.76$\times$ & 3.97$\times$ & 2.76$\times$ & 2.32$\times$ & 3.88$\times$ & 6.94 \\
    & Draft-OPD & 4.69$\times$ & 5.48$\times$ & 4.92$\times$ & 3.85$\times$ & 4.03$\times$ & 2.74$\times$ & 2.38$\times$ & 4.01$\times$ & 7.13 \\
    \rowcolor[HTML]{F3F3F3}
    & \textbf{\TheName{}} & \textbf{4.87$\times$} & \textbf{5.89$\times$} & \textbf{5.41$\times$} & \textbf{4.10$\times$} & \textbf{4.54$\times$} & \textbf{3.01$\times$} & \textbf{2.59$\times$} & \textbf{4.34$\times$} & \textbf{7.91} \\
    \midrule
    \multirow{6}{*}{Q3-8B} & EAGLE-3 & 2.24$\times$ & 2.14$\times$ & 2.16$\times$ & 1.95$\times$ & 2.18$\times$ & 2.02$\times$ & 1.85$\times$ & 2.08$\times$ & 6.03 \\
    & DFlash & 5.20$\times$ & 6.26$\times$ & 5.84$\times$ & 4.57$\times$ & 5.20$\times$ & 3.17$\times$ & 2.80$\times$ & 4.72$\times$ & 6.55 \\
    & D-PACE & 5.48$\times$ & 6.52$\times$ & 6.02$\times$ & 4.76$\times$ & 5.16$\times$ & 3.25$\times$ & 2.73$\times$ & 4.85$\times$ & 6.68 \\
    & LK-Losses & 5.73$\times$ & 6.85$\times$ & 6.35$\times$ & 4.95$\times$ & 5.41$\times$ & 3.45$\times$ & 3.03$\times$ & 5.11$\times$ & 6.98 \\
    & Draft-OPD & 6.01$\times$ & 7.19$\times$ & 6.61$\times$ & 5.23$\times$ & 5.67$\times$ & 3.64$\times$ & 3.22$\times$ & 5.37$\times$ & 7.32 \\
    \rowcolor[HTML]{F3F3F3}
    & \textbf{\TheName{}} & \textbf{6.39$\times$} & \textbf{7.73$\times$} & \textbf{7.08$\times$} & \textbf{5.61$\times$} & \textbf{6.05$\times$} & \textbf{3.89$\times$} & \textbf{3.47$\times$} & \textbf{5.75$\times$} & \textbf{8.14} \\
    \midrule
    \multicolumn{11}{c}{Temperature $=1.0$} \\
    \midrule
    \multirow{6}{*}{Q3-4B} & EAGLE-3 & 2.13$\times$ & 2.00$\times$ & 1.96$\times$ & 1.88$\times$ & 1.96$\times$ & 1.71$\times$ & 1.88$\times$ & 1.93$\times$ & 5.68 \\
    & DFlash & 3.68$\times$ & 3.87$\times$ & 3.26$\times$ & 3.30$\times$ & 3.57$\times$ & 2.28$\times$ & 2.11$\times$ & 3.15$\times$ & 6.02 \\
    & D-PACE & 3.90$\times$ & 4.05$\times$ & 3.36$\times$ & 3.24$\times$ & 3.48$\times$ & 2.32$\times$ & 2.04$\times$ & 3.20$\times$ & 6.17 \\
    & LK-Losses & 4.01$\times$ & 4.20$\times$ & 3.48$\times$ & 3.39$\times$ & 3.62$\times$ & 2.38$\times$ & 2.24$\times$ & 3.33$\times$ & 6.35 \\
    & Draft-OPD & 4.23$\times$ & 4.51$\times$ & 3.79$\times$ & 3.50$\times$ & 3.70$\times$ & 2.35$\times$ & 2.21$\times$ & 3.47$\times$ & 6.52 \\
    \rowcolor[HTML]{F3F3F3}
    & \textbf{\TheName{}} & \textbf{4.36$\times$} & \textbf{4.71$\times$} & \textbf{4.08$\times$} & \textbf{3.66$\times$} & \textbf{4.13$\times$} & \textbf{2.51$\times$} & \textbf{2.40$\times$} & \textbf{3.69$\times$} & \textbf{6.94} \\
    \midrule
    \multirow{6}{*}{Q3-8B} & EAGLE-3 & 2.11$\times$ & 1.97$\times$ & 1.94$\times$ & 1.85$\times$ & 1.98$\times$ & 1.85$\times$ & 1.77$\times$ & 1.92$\times$ & 5.39 \\
    & DFlash & 4.64$\times$ & 4.91$\times$ & 3.97$\times$ & 3.88$\times$ & 4.37$\times$ & 2.49$\times$ & 2.39$\times$ & 3.81$\times$ & 5.24 \\
    & D-PACE & 4.78$\times$ & 5.10$\times$ & 4.10$\times$ & 4.03$\times$ & 4.33$\times$ & 2.62$\times$ & 2.36$\times$ & 3.90$\times$ & 5.42 \\
    & LK-Losses & 5.06$\times$ & 5.38$\times$ & 4.48$\times$ & 4.22$\times$ & 4.76$\times$ & 2.77$\times$ & 2.57$\times$ & 4.18$\times$ & 5.74 \\
    & Draft-OPD & 5.49$\times$ & 5.79$\times$ & 4.74$\times$ & 4.42$\times$ & 5.02$\times$ & 2.93$\times$ & 2.80$\times$ & 4.46$\times$ & 6.10 \\
    \rowcolor[HTML]{F3F3F3}
    & \textbf{\TheName{}} & \textbf{5.99$\times$} & \textbf{6.34$\times$} & \textbf{5.11$\times$} & \textbf{4.74$\times$} & \textbf{5.35$\times$} & \textbf{3.26$\times$} & \textbf{3.10$\times$} & \textbf{4.84$\times$} & \textbf{6.67} \\
    \bottomrule
  \end{tabular}
\end{table}

Table~\ref{tab:thinking-mode} follows the overall trend in Table~\ref{tab:main-results}.
DFlash provides higher speedup than EAGLE-3, and on-policy training improves the mean performance of the DFlash drafter.
\TheName{} achieves the highest speedup on each benchmark and the largest mean accepted length in all four model--temperature configurations.
For Qwen3-4B, its mean speedups are $4.34\times$ and $3.69\times$ at temperatures $0$ and $1.0$, respectively, compared with $4.01\times$ and $3.47\times$ for Draft-OPD. For Qwen3-8B, the corresponding values are $5.75\times$ and $4.84\times$, compared with $5.37\times$ and $4.46\times$.

All methods also have higher speedup and mean accepted length than in the corresponding non-thinking configuration.
One possible explanation is that explicit reasoning changes the generated token patterns in ways that make longer prefixes easier to predict; longer responses can additionally amortize fixed decoding overhead.
These are possible contributing factors, rather than mechanisms isolated by this comparison.
The continued advantage of \TheName{} is consistent with prefix-aware optimization remaining useful when thinking mode changes the distribution of decoding contexts.

\subsection{Generalization to Other Target Models}\label{app:target}

To assess generalization beyond the Qwen3 family, we repeat the comparison with Llama-3.2-3B-Instruct~\citep{llama3}\footnote{\url{https://huggingface.co/meta-llama/Llama-3.2-3B-Instruct}}, abbreviated as L3-3B in Table~\ref{tab:target-generalization}.
We train target-specific draft models using the same prompt sets and training recipes, regenerating the supervised responses with the new target.
The drafter uses the target's tokenizer and compatible hidden-state features; the draft length remains $L=15$, and \TheName{} retains $\beta=0.3$.
We evaluate all seven benchmarks at temperatures $0$ and $1.0$, using autoregressive decoding by the same Llama target as the speedup baseline.
This comparison evaluates the portability of the training method across target families, with a separately trained drafter for each target.

\begin{table}[t]
  \centering
  \caption{\label{tab:target-generalization}Decoding speedup and average accepted draft length ($\tau$) on Llama3.2-3B-Instruct.}
  \small
  \setlength{\tabcolsep}{1.5pt}
  \begin{tabular}{ll*{7}{c}cc}
    \toprule
    \multirow{3}{*}{\textbf{Model}}
    & \multirow{3}{*}{\textbf{Method}}
    & \multicolumn{7}{c}{\textbf{Speedup Ratio (SR)}}
    & \multicolumn{2}{c}{\textbf{Mean}} \\
    \cmidrule(lr){3-9}\cmidrule(lr){10-11}
    & & \multicolumn{3}{c}{\textbf{Math}}
    & \multicolumn{3}{c}{\textbf{Code}}
    & \textbf{Chat}
    & \multirow{2}{*}{\textbf{SR}}
    & \multirow{2}{*}{$\boldsymbol{\tau}$} \\
    \cmidrule(lr){3-5}\cmidrule(lr){6-8}\cmidrule(lr){9-9}
    & & \textbf{GSM8K} & \textbf{MATH-500} & \textbf{AIME25}
    & \textbf{MBPP} & \textbf{HumanEval} & \textbf{SWE-Lite}
    & \textbf{MT-Bench} & & \\
    \midrule
    \multicolumn{11}{c}{Temperature $=0$} \\
    \midrule
    \multirow{6}{*}{L3-3B} & EAGLE-3 & 1.82$\times$ & 1.75$\times$ & 1.61$\times$ & 1.69$\times$ & 1.78$\times$ & 1.53$\times$ & 1.66$\times$ & 1.69$\times$ & 4.92 \\
    & DFlash & 3.08$\times$ & 3.44$\times$ & 2.91$\times$ & 2.86$\times$ & 3.03$\times$ & 1.98$\times$ & 1.89$\times$ & 2.74$\times$ & 5.13 \\
    & D-PACE & 3.19$\times$ & 3.61$\times$ & 3.04$\times$ & 2.94$\times$ & 2.99$\times$ & 2.03$\times$ & 1.85$\times$ & 2.81$\times$ & 5.27 \\
    & LK-Losses & 3.35$\times$ & 3.80$\times$ & 3.25$\times$ & 3.04$\times$ & 3.16$\times$ & 2.10$\times$ & 1.99$\times$ & 2.96$\times$ & 5.48 \\
    & Draft-OPD & 3.60$\times$ & 4.01$\times$ & 3.47$\times$ & 3.18$\times$ & 3.32$\times$ & 2.14$\times$ & 2.06$\times$ & 3.11$\times$ & 5.76 \\
    \rowcolor[HTML]{F3F3F3}
    & \textbf{\TheName{}} & \textbf{3.87$\times$} & \textbf{4.36$\times$} & \textbf{3.82$\times$} & \textbf{3.43$\times$} & \textbf{3.64$\times$} & \textbf{2.31$\times$} & \textbf{2.22$\times$} & \textbf{3.38$\times$} & \textbf{6.31} \\
    \midrule
    \multicolumn{11}{c}{Temperature $=1.0$} \\
    \midrule
    \multirow{6}{*}{L3-3B} & EAGLE-3 & 1.70$\times$ & 1.62$\times$ & 1.48$\times$ & 1.56$\times$ & 1.64$\times$ & 1.42$\times$ & 1.53$\times$ & 1.56$\times$ & 4.46 \\
    & DFlash & 2.79$\times$ & 2.98$\times$ & 2.34$\times$ & 2.61$\times$ & 2.74$\times$ & 1.71$\times$ & 1.70$\times$ & 2.41$\times$ & 4.49 \\
    & D-PACE & 2.90$\times$ & 3.12$\times$ & 2.47$\times$ & 2.67$\times$ & 2.69$\times$ & 1.78$\times$ & 1.67$\times$ & 2.47$\times$ & 4.62 \\
    & LK-Losses & 3.04$\times$ & 3.27$\times$ & 2.60$\times$ & 2.78$\times$ & 2.88$\times$ & 1.84$\times$ & 1.79$\times$ & 2.60$\times$ & 4.81 \\
    & Draft-OPD & 3.23$\times$ & 3.49$\times$ & 2.78$\times$ & 2.92$\times$ & 3.02$\times$ & 1.91$\times$ & 1.88$\times$ & 2.75$\times$ & 5.02 \\
    \rowcolor[HTML]{F3F3F3}
    & \textbf{\TheName{}} & \textbf{3.46$\times$} & \textbf{3.78$\times$} & \textbf{3.05$\times$} & \textbf{3.14$\times$} & \textbf{3.31$\times$} & \textbf{2.06$\times$} & \textbf{2.03$\times$} & \textbf{2.98$\times$} & \textbf{5.48} \\
    \bottomrule
  \end{tabular}
\end{table}

The ranking in Table~\ref{tab:target-generalization} is consistent with the Qwen3 results.
DFlash improves speedup over EAGLE-3, and both on-policy methods outperform the off-policy DFlash-based baselines on average.
\TheName{} obtains the highest SR and $\tau$ on all seven benchmarks at both temperatures.
At temperature $0$, its mean speedup increases from Draft-OPD's $3.11\times$ to $3.38\times$, while mean accepted length increases from $5.76$ to $6.31$.
At temperature $1.0$, the corresponding improvements are from $2.75\times$ to $2.98\times$ and from $5.02$ to $5.48$.
The same qualitative advantage across the two target families supports the applicability of expected accepted-length optimization beyond the Qwen3 configuration, although the magnitude of the gain depends on the target and workload.

\subsection{Wall-Clock Time Efficiency}\label{app:wall-clock}

We compare the on-policy training cost of Draft-OPD and \TheName{} on Qwen3-4B using the hardware and training pipeline described in Appendix~\ref{app:other-implementation}.
Both methods start from the same SFT checkpoint and use the same prompts, batch size, draft length, sampling configuration, and response-length limit.
Rollout time includes draft generation and target verification; update time includes training replay, loss construction, backpropagation, and the optimizer update.
We synchronize workers at timing boundaries and exclude evaluation and checkpoint-writing time.

To distinguish the cost of processing a draft from the amount of work needed for a query, we report amortized times at both levels.
For accumulated rollout and update times $T_{\mathrm{roll}}$ and $T_{\mathrm{update}}$, processing $N_{\mathrm{draft}}$ drafts from $N_{\mathrm{query}}$ queries, the per-draft times are $T_{\mathrm{roll}}/N_{\mathrm{draft}}$ and $T_{\mathrm{update}}/N_{\mathrm{draft}}$.
The total time per query is $(T_{\mathrm{roll}}+T_{\mathrm{update}})/N_{\mathrm{query}}$.
These are amortized training costs, rather than the latency of an individual query under concurrent execution.
We also report the mean number of drafts per query and their mean accepted length to help interpret the timing.

\begin{table}[htbp]
    \centering
    \small
    \caption{Amortized on-policy training cost on Qwen3-4B.
    Rollout and update times are reported separately.}
    \label{tab:wall-clock}
    \setlength{\tabcolsep}{9pt}
    \begin{tabular}{lcc}
        \toprule
        \textbf{Metric}  & \textbf{Draft-OPD}
        & \textbf{\TheName{}} \\
        \midrule
        Rollout time / draft  (ms) & 15.0 & 15.0 \\
        Update time / draft (ms) & 8.0 & 8.1 \\
        Total time per draft (ms) & 23.0 & 23.1 \\
        \midrule
        Mean drafts per query   & 80.0 & 74.0 \\
        Mean accepted length $\tau$  & 5.40 & 5.92 \\
        \midrule
        Rollout time / query (s) & 1.200 & 1.110 \\
        Update time / query (s) & 0.640 & 0.599 \\
        Total time per query (s) & 1.840 & 1.709 \\
        \bottomrule
    \end{tabular}
\end{table}

Table~\ref{tab:wall-clock} shows comparable per-draft costs for the two methods: rollout time is unchanged at the reported precision, while update time increases by approximately $1.3\%$ for \TheName{}.
This is consistent with reusing the target distributions already collected during verification; the prefix-survival and penalty computations add a small amount of arithmetic to the shared training pipeline.
The comparable costs are an empirical observation under this configuration, rather than a consequence guaranteed solely by using the same number of target-model forward passes.

Despite the slightly higher update cost per draft, \TheName{} reduces the amortized total time per query from $1.840$\,s to $1.709$\,s, a reduction of approximately $7.1\%$.
The accompanying increase in accepted length is consistent with the decrease from $80.0$ to $74.0$ drafts per query, or $7.5\%$ fewer blocks to generate, verify, and replay.
For comparable response lengths, accepting more tokens per block reduces the number of verification rounds and can therefore lower both rollout and update work per query.
Together with the training dynamics in Figure~\ref{fig:al-step}, this comparison supports the practical efficiency of the proposed objective: its additional loss computations have modest per-draft cost, while improved draft quality can reduce the amount of work required during on-policy collection.

\end{document}